\documentclass[11pt]{article}

\usepackage[preprint]{acl}
\usepackage{times}
\usepackage{latexsym}

\usepackage[T1]{fontenc}

\usepackage[utf8]{inputenc}
\usepackage{microtype}
\usepackage{inconsolata}
\usepackage{graphicx}
\usepackage{todonotes}
\usepackage{booktabs}
\usepackage{amsthm}
\usepackage{amsmath}
\usepackage{amsfonts}
\usepackage{multirow}
\usepackage[svgnames]{xcolor}
\usepackage{xspace}
\usepackage[listings]{tcolorbox}
\tcbuselibrary{breakable}
\usepackage{paralist}
\usepackage{float}
\usepackage{cleveref}

\newcommand{\flores}{Flores+\xspace}
\newcommand{\langdir}[2]{\textit{#1}$\rightarrow$\textit{#2}\xspace}
\newcommand{\pmsx}{\langdir{Pms}{X}}
\newcommand{\xpms}{\langdir{X}{Pms}}

\newcommand{\llama}{Llama\xspace}
\newcommand{\gemma}{Gemma\xspace}
\newcommand{\qwen}{Qwen\xspace}
\newcommand{\eurollm}{EuroLLM\xspace}
\newcommand{\tower}{Tower\xspace}
\newcommand{\gemini}{Gemini\xspace}
\newcommand{\gpt}{GPT\xspace}
\newcommand{\chrf}{chrF++\xspace}
\newcommand{\fscore}{F$_{1}$\xspace}
\newcommand{\ua}{$\uparrow$\xspace}

\title{Crowdsourcing Piedmontese to Test LLMs on Non-Standard Orthography}

\author{Gianluca Vico \and Jindřich Libovický \\
        Charles University, Faculty of Mathematics and Physics, \\
        Institute of Formal and Applied Linguistics \\
        Malostranské náměstí 25, 118 00 Praha, Czech Republic\\
        \texttt{\{vico, libovicky\}}@ufal.mff.cuni.cz}

\begin{document}
\maketitle
\begin{abstract}

We present a crowdsourced dataset for Piedmontese, an endangered Romance language of northwestern Italy. The dataset comprises 145 Italian–Piedmontese parallel sentences derived from \flores, with translations produced by speakers writing in their natural orthographic style rather than adhering to standardized conventions, along with manual word alignment. We use this resource to benchmark several large language models on tokenization parity, topic classification, and machine translation. Our analysis reveals that Piedmontese incurs a tokenization penalty relative to higher-resource Romance languages, yet LLMs achieve classification performance approaching that of Italian, French, and English. Machine translation results are asymmetric: models translate adequately from Piedmontese into high-resource languages, but generation into Piedmontese remains challenging. The dataset and code are publicly released.

\end{abstract}

\section{Introduction}
Piedmontese (ISO 639-3: \texttt{pms}) is a Romance language spoken in the Piedmont region of northwestern Italy. According to Ethnologue \citep{ethnologue28}, it has fewer than one million speakers and is classified as endangered, with intergenerational transmission in decline.

Existing NLP resources for Piedmontese are limited and predominantly derived from Piedmontese Wikipedia. While useful, these sources largely adhere to standardized orthographic conventions and thus fail to capture the orthographic variations that are common in written Piedmontese. This discrepancy raises the question of how well current language models handle Piedmontese as it is actually written by speakers.

\begin{table}[t]
    \centering
    \begin{tabular}{lp{0.8\linewidth}}
        \toprule
        \multicolumn{2}{c}{\flores \textit{dev} 114} \\
        \midrule
        Ita & Si tratta della maggiore acquisizione nella storia di eBay.\\
        Pms & A l'è la pi granda aquisission ënt la stòria d'ebay \\
        Eng & It is the biggest acquisition in eBay's history.\\
        Fra & C’est la plus grande acquisition de l’histoire d’eBay. \\
        \bottomrule
    \end{tabular}
    \caption{Sample from parallel sentences for evaluating machine translation. Annotators translated the Italian sample into Piedmontese. The Italian, French and English samples are originally from \flores.}
    \label{tab:parallel-sample}
\end{table}

To address this gap, we present a crowdsourced dataset of Italian-to-Piedmontese translations, where annotators were explicitly instructed to write in whichever orthographic style feels natural to them. The source sentences are drawn from the \flores dataset \citep{flores}, a multiparallel corpus spanning over 200 languages. We additionally provide manual word alignments between Piedmontese and Italian sentence pairs.

Using this data, we evaluate several large language models (LLMs) both intrinsically, through tokenization parity \cite{parity} analysis, and extrinsically, on topic classification (using labels from SIB-200; \citealp{sib}) and machine translation (MT). Our results indicate that current LLMs exhibit reasonable comprehension of Piedmontese, achieving decent performance in classification and translation from Piedmontese into high-resource languages. 
However generation into Piedmontese remains substantially more challenging.

We illustrate the data collection procedure in Section~\ref{sec:collection}. In Section~\ref{sec:dataset}, we describe the dataset, and in Section~\ref{sec:eval}, we asses LLM performance on Piedmontese. Section~\ref{sec:related} presents related datasets and in Section~\ref{sec:conclusion} we summarise our findings.

The dataset\footnote{\url{http://hdl.handle.net/11372/LRT-6086}} and evaluation code\footnote{\url{https://github.com/GianlucaVico/CrowdsourcedPiedmontese}} are released under an open-source license.

\begin{figure*}[t]
    \centering
    \includegraphics[width=0.9\linewidth]{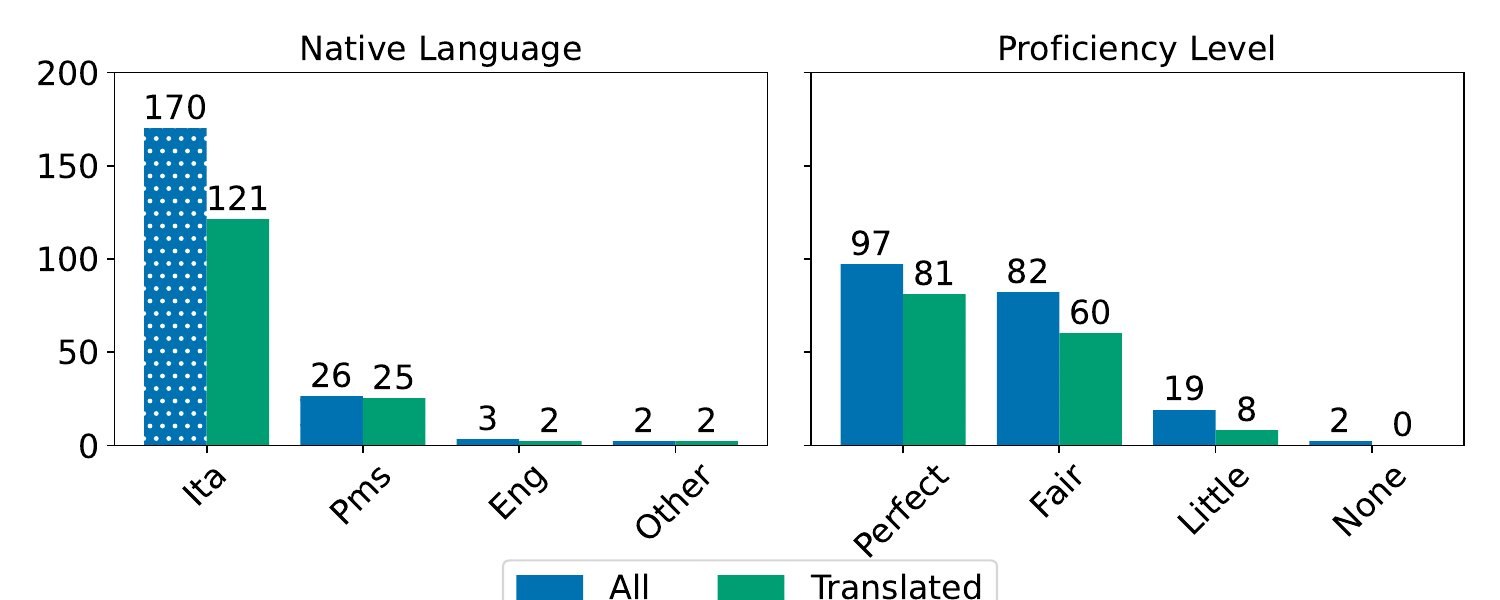}
    \caption{On the left, the main language used by the annotators; Icelandic is included in \textit{Other}. On the right, the self-reported proficiency in Piedmontese. The majority of people uses Italian and self-reports perfect or fair proficiency in Piedmontese.}
    \label{fig:language}
\end{figure*}

\begin{figure}[th]
    \centering
    \includegraphics[width=1.0\linewidth]{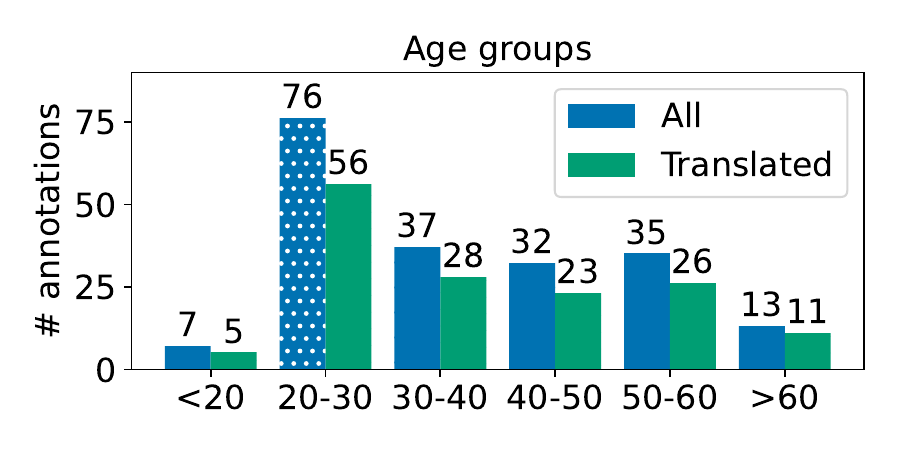}
    \caption{Age distribution of the annotators. Most annotators are 20-30 years old, while older people are more likely to know Piedmontese, but we did not reach them.}
    \label{fig:age}
\end{figure}

\section{Data Collection}\label{sec:collection}

We collected translations via an online questionnaire administered in Italian (see Appendix~\ref{app:question}), the dominant language in the region, understood by all Piedmontese speakers. Annotators were recruited voluntarily through social media and word of mouth, with no restrictions on repeated participation. To preserve anonymity, we did not track annotator identity across sessions.

The questionnaire comprises three components. First, we elicit demographic and sociolinguistic information, including the annotator's primary language, self-assessed proficiency in Piedmontese, age group, and method of language acquisition. We also ask whether they believe Piedmontese has a standard orthography and, if so, whether it is commonly used. These questions serve both to characterize our annotator population and to contextualize the orthographic variation in the resulting translations.

Second, we present annotators with a randomly selected Italian sentence from \flores \citep{flores} and ask them to translate it into Piedmontese. Crucially, annotators are instructed to write in whatever manner feels natural to them, rather than adhering to any prescribed standard. Translation is optional, so annotators can still complete the other parts of the questionnaire. In this way, we can accommodate annotators who comprehend Piedmontese but do not actively write it. To address the absence of certain diacritics on standard physical keyboards, we provide a substitution scheme (e.g., \textit{/:a} for \textit{\"a}).
Mobile keyboards do not have this issue, and we observed that people either directly use diacritics or use diacritics that can be found on Italian keyboards (\textit{àèéìòù}).

Third, annotators evaluate a translation submitted by a previous participant, viewing both the Italian source and the Piedmontese rendering. This peer review mechanism enables filtering of erroneous or inappropriate submissions and provides an estimate of inter-annotator agreement on translation quality. While the task is subjective, we ask annotators to take into consideration possible variations of the language and of the orthography. 

\begin{figure*}[th!]
    \centering
    \includegraphics[width=01\linewidth]{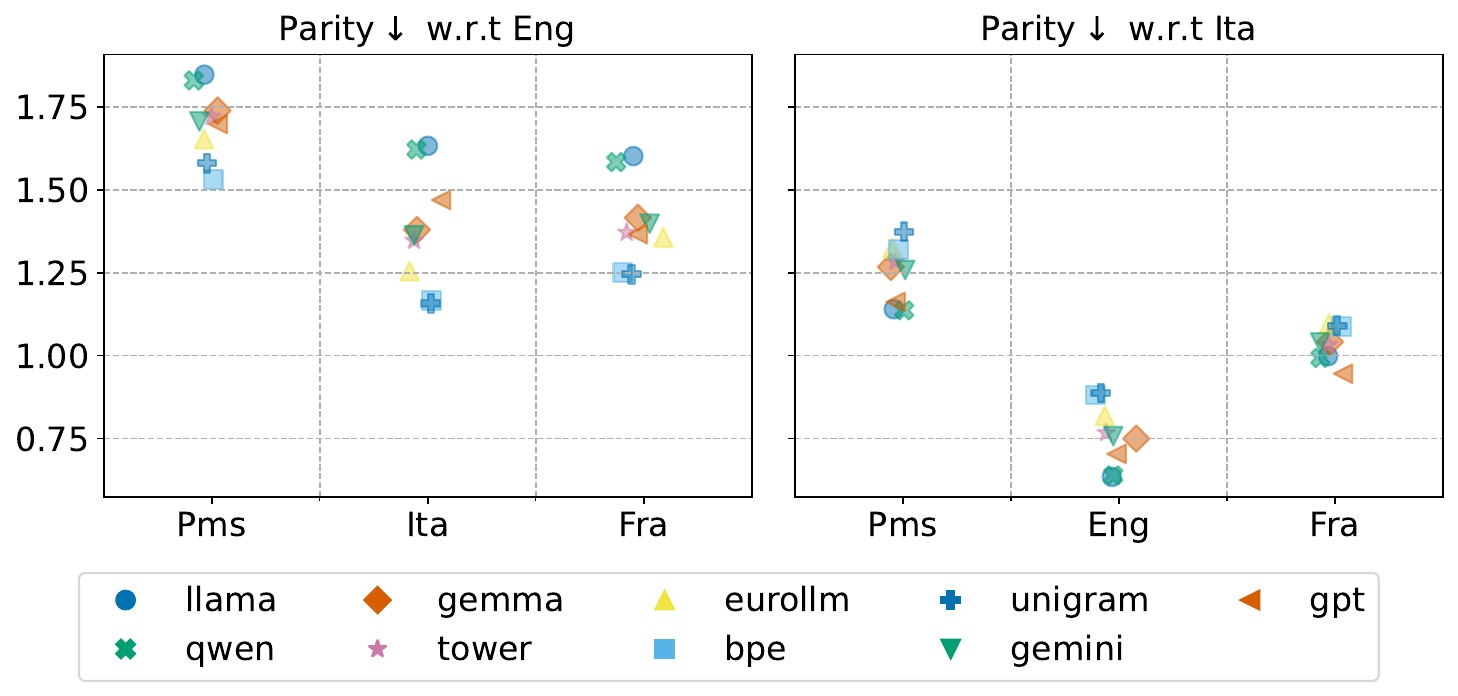}
    \caption{Parity scores with respect to English and Italian. Piedmontese has worse parity compared to the other languages; however, it is closer to one when compared to Italian.}
    \label{fig:parity}
\end{figure*}

\section{Dataset Description}\label{sec:dataset}

We have collected 200 annotations, and 145 of them have valid translations: 68 are from the \flores \textit{dev} set, while 77 are from the \textit{devtest} set. 102 samples have been reviewed by at least one annotator, but due to their limited number. We use the reviews only to filter missing or offensive translations.

We organise the collected data in three datasets: 1) the raw list of annotations that can be used for further analysis, 2) a list of parallel sentences for evaluating MT systems, and 3) a list of word-aligned sentences.

\subsection{Annotation}

Figure~\ref{fig:language} shows that most annotators use primarily Italian, and a few use Piedmontese. Other languages include English and Icelandic. The proportion of annotators who submitted a translation is higher among Piedmontese speakers than among Italian speakers.
Additionally, most annotators declared themselves to be perfectly or fully proficient in Piedmontese.
Most of the annotators are confident in their language knowledge; however, only a small portion considers Piedmontese their native language.

Our questionnaire reached mainly younger people, as shown in Figure~\ref{fig:age}, but, since this is an endangered language, older people are more likely to speak it.

On average, completing the questionnaire took approximately 7 minutes. People who did not provide a translation took approximately 3 minutes.
According to 11\% of the annotators, people use standard grammar when writing Piedmontese, while 42\% of them disagree. 54\% of them think that Piedmontese has a standard grammar, whether it is used or not, and for 25\% of the annotators, there is no standard.

\subsection{Parallel Sentences}
\flores contains 2009 samples divided into the \textit{dev} and \textit{devtest} splits.
The sentences provided to the annotators are randomly selected, so some of them have multiple translations: three samples from the \textit{devtest} set and one sample from the \textit{dev} set have two translations.
The paired sentences have the same overall meaning, but translation quality varies; for example, annotators may use more general terms, summarise a list or remove details.
102 samples have at least one human review, which we used to remove incorrect translations.
We present a sample in Table~\ref{tab:parallel-sample}: as can be seen, the Piedmontese text may contain incorrect capitalization or missing punctuation. Also, the use of diacritics is inconsistent among annotators.

\subsection{Word Aligned Sentences}

Due to the limited number of samples, the authors are able to manually word-align the Piedmontese and Italian sentences. 
We select pairs of corresponding spans in the paired sentences, ensuring that each span is non-overlapping (i.e., each word belongs to at most one span)
In this sense, there are cases where, for example, a verb is aligned with a noun because they convey the same meaning, but the sentence structure differs. We use the white space and apostrophe to split words. As an example, \textit{e sull'albero (and on the tree)} consists of three words: \textit{[e][sull'][albero]}.
One word can be aligned to multiple words, e.g., \textit{è (is)} is aligned to \textit{a l'è}, and there are unaligned words. 
However, we do not consider subword alignment.
The dataset comprises 3003 spans, with a median of 20 spans per sentence pair. 2902 spans are a single word aligned to another single word. 
The median number of characters for each span is 5 for both Italian and Piedmontese.

\section{Model Evaluation}\label{sec:eval}

To assess LLM performance on Piedmontese, we first evaluate tokenizer parity \cite{parity}: this provides an estimate of the costs in tokens of processing Piedmontese compared to other languages.
Then, we use the aligned dataset to investigate whether models can find corresponding words between Piedmontese and Italian. Finally, we use topic classification and machine translation as downstream tasks for evaluation. In topic classification, models need to be able to understand Piedmontese, while in machine translation, they also have to generate Piedmontese.
The downstream tasks are evaluated in a zero-shot setup.

We consider the following open-weight models from HuggingFace: \llama 3.3 70B\footnote{meta-llama/Llama-3.3-70B-Instruct} \cite{llama}, \gemma 3 27B\footnote{google/gemma-3-27b-it} \cite{gemma}, \qwen 3 30B\footnote{Qwen/Qwen3-30B-A3B-Instruct-2507} \cite{qwen}, \eurollm 9B\footnote{utter-project/EuroLLM-9B-Instruct} \cite{eurollm}, \tower Plus 9B\footnote{Unbabel/Tower-Plus-9B} \cite{tower}; and the closed-source models: \gemini\footnote{gemini-2.5-flash-preview-09-2025} and \gpt\footnote{gpt-4o-mini}.
Besides Piedmontese and Italian, we also include French, as it is the high-resource language closest to Piedmontese, other than Italian, and English, due to its widespread availability.
Because the data is limited and we do not perform any parameter search, we evaluate the model on the combined \textit{dev} and \textit{devtest} sets. The hyper-parameters for the experiments are listed in Appendix~\ref{app:hyp}.

\subsection{Tokenizer Parity}

As shown by \citet{ahia-etal-2023-languages}, low-resource languages are often overtokenized, resulting in higher costs and worse performance compared to high-resource languages.
We evaluate the tokenizer parity \cite{parity} for the LLMs, UnigramLM, and BPE from SentencePiece to estimate the number of extra tokens required to process the same sentence in Piedmontese.

We train the SentencePiece tokenizers using English, Italian, French, and Piedmontese data from the Glot500 Corpus \cite{imanigooghari-etal-2023-glot500}, with 100k samples for each language and a vocabulary size of 32,000.

We average the parity of each sample, computed as:
\begin{equation*}
    p_t(s_{\text{tgt}}, s_{\text{ref}}) = \dfrac{\vert t(s_{\text{tgt}})\vert}{\vert t(s_{\text{ref}})\vert}
\end{equation*}
where $t$ is a tokenization function that produces a list of tokens, $s_{\text{tgt}}$ is a sentence in the target language, and $s_{\text{ref}}$ is the corresponding sentence in the reference language.
As reference languages, we use English and Italian.
A parity close to one indicates that the tokenizer produces a similar number of tokens for the source and target languages, whereas values greater than one indicate that the target language is over-tokenized.

In Figure~\ref{fig:parity}, we report the parity scores of the models. Piedmontese has worse (i.e., higher) parity than the other languages, which means that using LLMs with Piedmontese is more computationally expensive. 
Training the tokenizer on Piedmontese can help, as BPE and UnigramLM have lower parity compared to English. However, overall, the models yield comparable results, and closed models do not have an advantage over the open-weight models. In Appendix~\ref{app:parity}, we report the exact parity scores for the different setups.

\begin{figure}[t]
    \centering
    \includegraphics[width=1\linewidth]{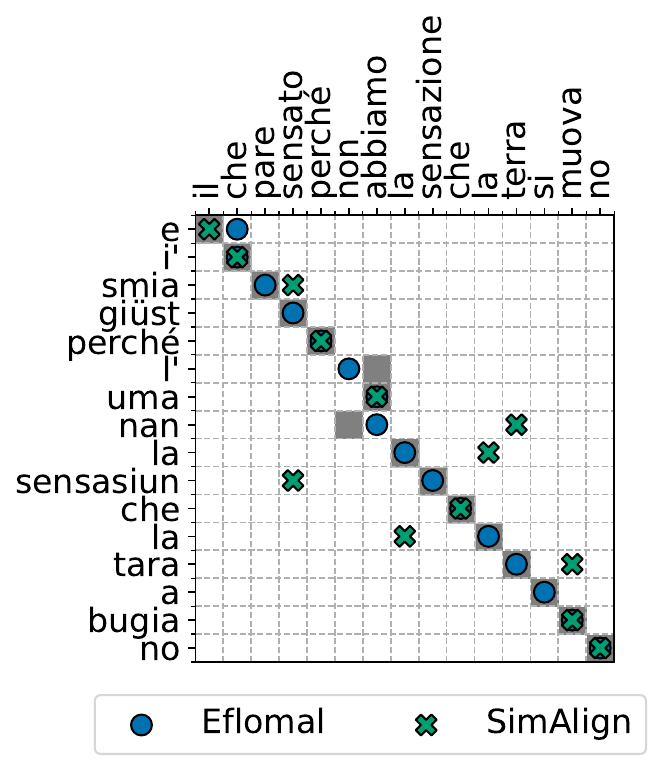}
    \caption{This is a sample alignment. The gray background is the reference alignment, while eflomal alignment is represented by the blue circles and SimAlign one by the green crosses. The English translation of the sentence is \textit{This seems sensible, because the Earth doesn't feel as if it's moving, does it?}}
    \label{fig:alignment}
\end{figure}
\begin{table}[t]
    \centering
    \begin{tabular}{lcccc}
        \toprule    
        Model & \fscore\ua & Precision\ua & Recall\ua\\
        \midrule            
        Eflomal  & .774 & .817 & .735\\
        SimAlign & .589 & .726 & .496\\    
        \bottomrule
    \end{tabular}
    \caption{Alignment scores of eflomal and SimAlign.}
    \label{tab:align}
\end{table}

\begin{figure*}[th]
    \centering
    \includegraphics[width=1\linewidth]{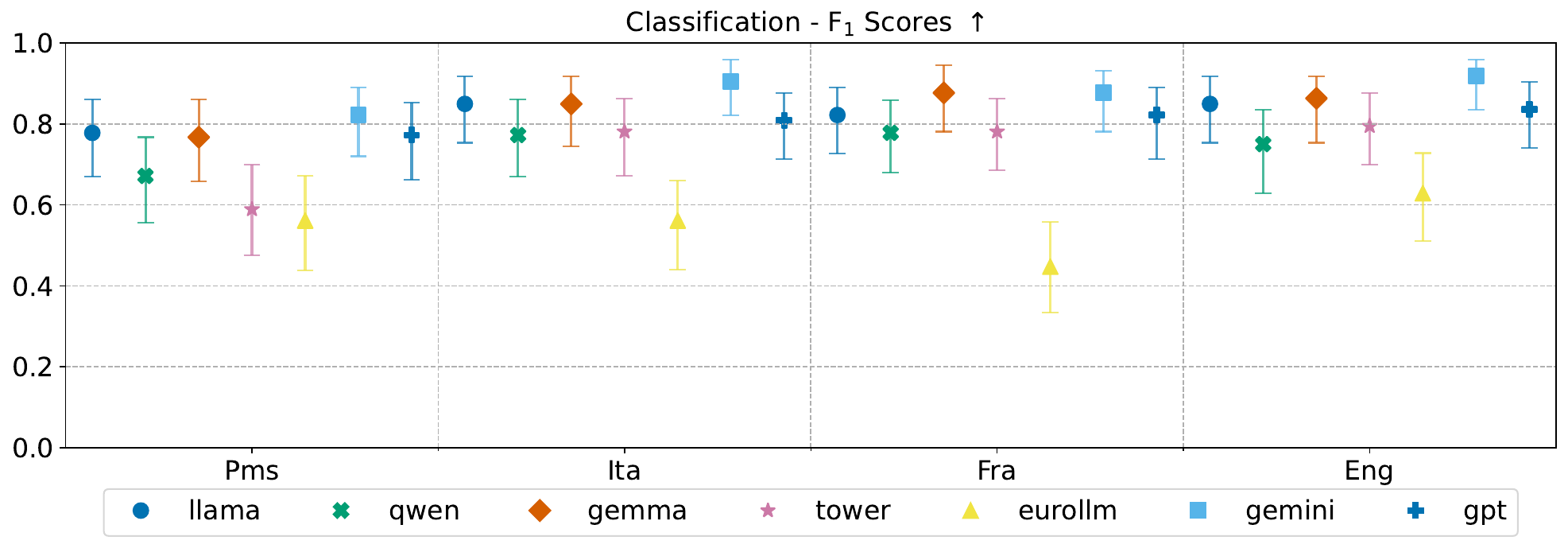}
    \caption{Comparison on the F1 scores of the models in the topic classification task. We perform bootstrapping to compute the confidence interval of the scores.}
    \label{fig:f1}
\end{figure*}

\begin{figure*}[th]
    \centering
    \includegraphics[width=1\linewidth]{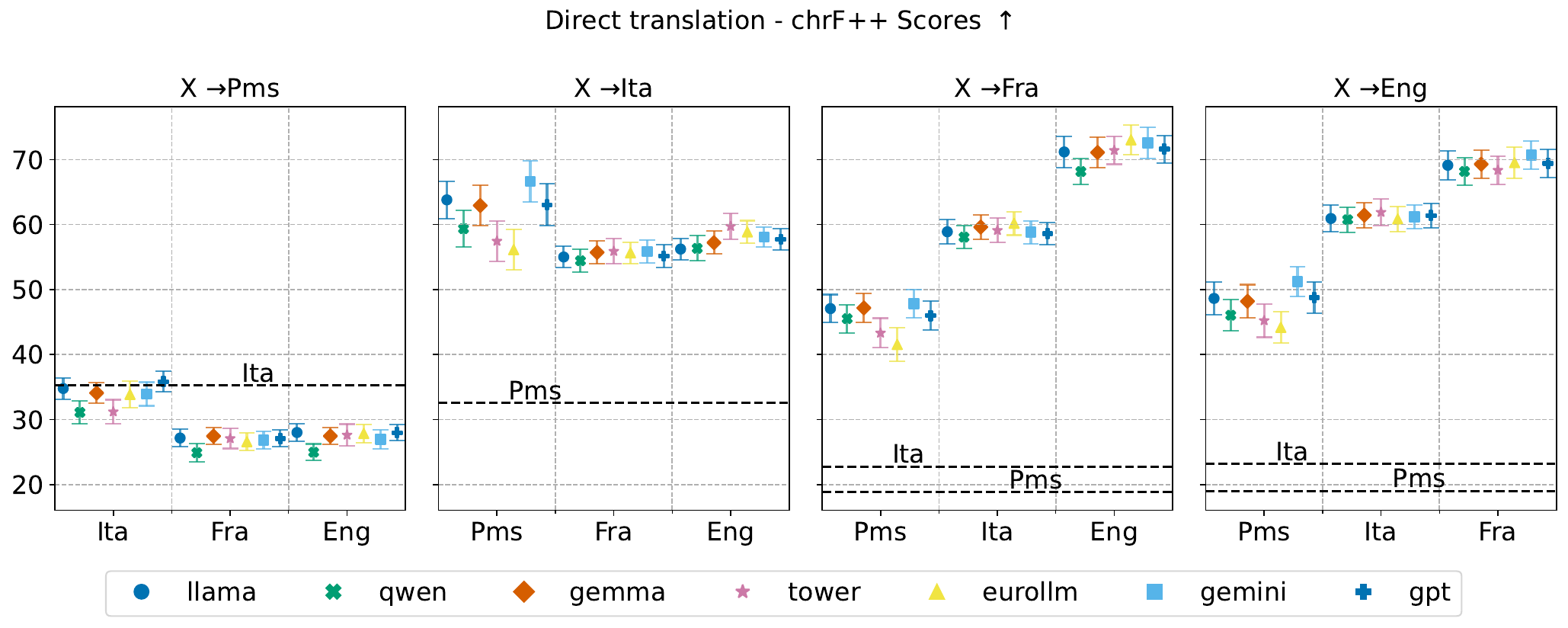}
    \caption{chrF++ scores of the different models. Each subplot shows the target language, while the sources languages are on the x-axis. The dotted horizontal lines indicate the scores obtained when using the reference text in a given language as if it were the translation.}
    \label{fig:chrf}
\end{figure*}

\subsection{Word Alignment}

We use eflomal \cite{eflomal} trained on our dataset as a baseline and compare with the unsupervised SimAlign \cite{simalign} with XLM-RoBERTa \cite{xlmr} with subwords. XLM-RoBERTa is a multilingual model, but Piedmontese was not explicitly included in the training data.
We use the same evaluation script from SimAlign, which reports precision, recall, \fscore, and alignment error rate (AER) defined as:
\begin{align*}
    &\text{prec} = \dfrac{\lvert A \cap P \vert}{\lvert A \rvert},\ 
    \text{rec} = \dfrac{\lvert A \cap S \vert}{\lvert S \rvert},\\
    &\text{F}_1 = \dfrac{2 \text{prec} \cdot \text{rec}}{\text{prec}+\text{rec}},\
    \text{AER} = 1 - \dfrac{\lvert A \cap S \rvert + \lvert A \cap P \vert}{\lvert A \rvert + \lvert S \rvert}    
\end{align*}
where $A$ are the system alignment, $S$ the sure reference alignment and $P$ the possible reference alignments. However, our annotations do not include possible alignment and so AER is simply $1-\text{F}_1$ (see Appendix~\ref{app:align-metric}). 

The results in Table~\ref{tab:align} show that eflomal achieves better scores than SimAlign, which relies on the language model representations. 
The scores of SimAlign are comparable to those that its authors observed for English-Hindi alignment, indicating that the model produces reasonable alignments despite not being trained on Piedmontese.
This indicates that while the XLM-RoBERTa representations are sufficient for generating zero-shot alignment, statistical methods still yield better results. 

Additionally, the effect of (sub)words that are identical between Italian and Piedmontese and are therefore easier to align is unknown.
We show an example of alignment in Figure~\ref{fig:alignment}, where the reference alignment is mostly monotonic. SimAlign seems to align words in the wrong position (e.g., \textit{la} aligned to the wrong \textit{la}), while eflomal might align words that occur together (e.g., the negation \textit{nan} with the verb \textit{abbiamo}).

\begin{table*}[th]
    \centering
    \begin{tabular}{cccccc}
        \toprule
         \multicolumn{2}{c}{Dev} & \multicolumn{4}{c}{Target}\\
         & & Pms & Ita & Fra & Eng \\
         \midrule
         \multirow{7}*{\rotatebox{90}{Source}} & Pms & - & 58.80 & 42.48 & 45.95 \\
         & Ita & 32.62 & - & 57.83 & 60.98 \\
         & Fra & 26.45 & 54.68 & - & 70.89 \\
         & Eng & 27.02 & 57.60 & 72.25 & - \\
         \cmidrule{2-6}
         & Pms\textsubscript{Pivot} & - & - & 44.62 & 46.91 \\
         & Fra\textsubscript{Pivot} & 27.67 & - & - & 67.71 \\
         & Eng\textsubscript{Pivot} & 28.07 & - & 67.96 & -\\
         \bottomrule
    \end{tabular}
    \
    \begin{tabular}{cccccc}
        \toprule
         \multicolumn{2}{c}{Devtest} & \multicolumn{4}{c}{Target}\\
         & & Pms & Ita & Fra & Eng \\
        \midrule
         \multirow{4}*{\rotatebox{90}{Source}} & Pms & - & 62.15 & 47.42 & 47.77 \\
         & Ita & 33.23 & - & 60.26 & 61.37 \\
         & Fra & 26.60 & 56.11 & - & 67.90 \\
         & Eng & 26.97 & 57.92 & 70.61 & -\\
         \cmidrule{2-6}
         & Pms\textsubscript{Pivot} & - & - & 49.57 & 49.91 \\
         & Fra\textsubscript{Pivot} & 27.11 & - & - & 65.11 \\
         & Eng\textsubscript{Pivot} & 27.85 & - & 67.63 & -\\
         \bottomrule
    \end{tabular}
    \caption{Average \chrf scores of the models on the two sets and all directions. Note that only columns are comparable. \textit{Pivot} refers to the experiments that use Italian as a pivot for the translation.}
    \label{tab:chrf}
\end{table*}

\begin{table}[t]
    \centering
    \begin{tabular}{p{0.2\linewidth}p{0.6\linewidth}}
        \toprule
        \multicolumn{2}{c}{\flores \textit{dev} 114}\\
        \multicolumn{2}{c}{\textit{It is the biggest acquisition in eBay's history.}} \\
        Pms & A l'è la pi granda aquisission ënt la stòria d'ebay \\
        Ita & Si tratta della maggiore acquisizione nella storia di eBay. \\
        \midrule
        \multicolumn{2}{c}{\langdir{Pms}{Ita}}\\
        \midrule
        EuroLLM & È la più grande acquisizione nella storia di eBay \\
        Gemini & È la più grande acquisizione nella storia di eBay \\
        \midrule
        \multicolumn{2}{c}{\langdir{Ita}{Pms}}\\
        \midrule
        Gemma & A l'é la pì gròssa acquisission an la stòria d'eBay. \\
        GPT & A l'é la piò gròssa aquisission an sla storia ëd eBay. \\
        \bottomrule
    \end{tabular}
    \caption{Translation examples. The \langdir{Ita}{Pms} translations are understandable, but have different spellings than the reference and across models. The \langdir{Pms}{Ita} translations are correct, although, phrased differently than the reference translation.}
    \label{tab:mt_examples}
\end{table}

\subsection{Topic Classification}

We use SIB-200 \cite{sib} to evaluate the models on topic classification with our data. SIB-200 uses sentences from \flores, so it is possible to obtain labels for the Piedmontese sentences.
The dataset contains 7 classes: \textit{science/technology}, \textit{travel}, \textit{politics}, \textit{sports}, \textit{health}, \textit{entertainment}, and \textit{geography}, but some sentences are labelled as \textit{uncategorized} and are excluded from our experiments.
In total, 37 sentences from the \textit{dev} set and 38 from the \textit{devtest} set have a label.
We use the same set of sentences for all four languages.

In Figure~\ref{fig:f1}, we report the \fscore scores of the models on the different languages. We note that, while scores for Piedmontese are generally lower than for the other languages, they are still comparable, meaning that models are able to understand the language.
Smaller models, such as \eurollm, exhibit worse performance: in particular, \eurollm struggles to follow instructions and, in French, often generates all labels or overly long explanations.
\tower has a larger drop in performance in Piedmontese, but it is still able to solve the task despite its focus on machine translation.
Closed models behave similarly to the larger open-weight models.
See Appendix~\ref{app:class} for the exact values and additional metrics and Appendix~\ref{app:promps} for the prompts used.

\subsection{Machine Translation}
We test zero-shot machine translation, including both \pmsx and \xpms.
From Figure~\ref{fig:chrf}, models have similar \chrf (\ua) scores when translating from the different languages to Piedmontese.  
Moreover, all languages achieve similar \chrf scores when translating to Italian, including Piedmontese. 
However, translating from Piedmontese to French or English is worse than in other languages.
While we cannot directly compare the target languages, \xpms has noticeably lower scores.

Given that \langdir{X}{ita} has comparable results for all languages, we use Italian as a pivot by first translating from the source language to Italian, and then to the target language.
From Table~\ref{tab:chrf}, the pivot strategy improves translations up to +2.15 \chrf in the \pmsx direction and +1.22 \chrf in the \xpms direction.

However, evaluating \xpms is particularly challenging, because models might produce Piedmontese that is correct but different from the reference, which does not use standard orthography, and surface-level metrics such as \chrf penalize this. Machine-learned metrics like COMET \cite{rei-etal-2020-comet} can improve this, but they need training data, which is not available.
In Appendix~\ref{app:mt}, we report additional metrics, including COMET (without fine-tuning on Piedmontese).
Moreover, we observe that the Italian sentences are closer to the Piedmontese references than what the models are generating, as shown by the horizontal lines in Figure~\ref{fig:chrf}. 
This can also explain why LLMs are able to understand Piedmontese. In Table~\ref{tab:mt_examples}, we show some translation examples between Italian and Piedmontese from different models.
See Appendix~\ref{app:mt} for the exact values and additional metrics and Appendix~\ref{app:promps} for the prompts used.

\section{Related Work}\label{sec:related}

The data for this work is derived from \flores \cite{flores}, which is an evaluation benchmark for machine translation. It contains 2009 sentences, each translated into more than 200 languages. It does not include Piedmontese, but it includes geographically close Italian regional languages, such as Ligurian and Lombard.
There are other datasets that contain Piedmontese data, such as Wikipedia (68k samples) and Wikisource (4k samples) \cite{wikidump}, and Glot500 \cite{imanigooghari-etal-2023-glot500} (226k samples), which derive from the Piedmontese portion of Wikipedia, and Tatoeba \cite{tatoeba} (800 samples), which contains sentences annotated by volunteers. However, these datasets contain a more standardised version of Piedmontese, which differs from what people might use in real life.

Datasets derived from CommonCrawl\footnote{\url{https://commoncrawl.org/}}, like C4 \cite{c4}, FineWeb2 \cite{fineweb2}, CulturaX \cite{culturax}, and Oscar \cite{oscar}, also contain some Piedmontese, but correctly identifying a low-resource language is challenging, and false positives can affect the results.

Another project with the objective of collecting language data, including Piedmontese, is AlpiLinK \cite{alpilink}. AlpiLinK collects crowdsourced spoken data of various regional languages in the Alpine regions of Italy and contains 5442 Piedmontese sentences.

\citet{diatopit} propose \textsc{DiatopIt}, a corpus of social media posts written in different local languages of Italy or using regional Italian. The corpus includes 288 Piedmontese samples and, similarly to our work, does not assume a standard orthography but focuses on the languages as actually written by people.

\section{Conclusion}\label{sec:conclusion}

In this paper, we presented a crowdsourced dataset for Piedmontese, whose main characteristic is the non-standard orthography. 
The dataset can be used for further research on the annotators' demographics, machine translation, and word alignment.
Furthermore, we highlight how Piedmontese is at a disadvantage in many popular NLP models, showing that it has higher parity compared to related languages.
We then test several LLMs to investigate their performance on topic classification and machine translation tasks. We note that models are able to understand Piedmontese, although they perform worse than in other languages; the scores are still comparable. However, generation still remains challenging. 

\section{Limitations}
This work presents several limitations. 
Firstly, the selection of annotators is biased because it relies on social media, and people who speak the language may not be accessible.
This also influences how the translations are written, because some characters are easier to type on a smartphone keyboard than on a physical one or with pen and paper.
Additionally, we do not track which variant of Piedmontese the annotators use, but we consider Piedmontese to be what the annotators themselves refer to as Piedmontese.
Then, the annotators are mostly Italian native speakers, since Italian is the national language, and the number of samples is extremely small. We focus on Piedmontese in Italy and do not consider, for example, Piedmontese spoken in Argentina. The task involves translating from Italian, which can result in translationese.
Also, in the questionnaire, we use the terms \textit{orthography} and \textit{grammar} interchangeably to make it easier to understand.

\section*{Acknowledgments}
This research was supported by the Czech Science Foundation project 25-16242S.
The work described herein has also been supported by the Ministry of Education, Youth and Sports of the Czech Republic, Project No. LM2023062 LINDAT/CLARIAH-CZ.
We thank the annotators who contributed to this work.
GV thanks friends and relatives and the Instagram pages piemontays, Spurgatocn and Abitare il Piemontese for sharing the questionnaire to a larger public.

\bibliography{custom,anthology-1,anthology-2}

\begin{thebibliography}{23}
\providecommand{\natexlab}[1]{#1}

\bibitem[{Adelani et~al.(2024)Adelani, Liu, Shen, Vassilyev, Alabi, Mao, Gao, and Lee}]{sib}
David~Ifeoluwa Adelani, Hannah Liu, Xiaoyu Shen, Nikita Vassilyev, Jesujoba~O. Alabi, Yanke Mao, Haonan Gao, and En-Shiun~Annie Lee. 2024.
\newblock \href {https://doi.org/10.18653/v1/2024.eacl-long.14} {{SIB}-200: A simple, inclusive, and big evaluation dataset for topic classification in 200+ languages and dialects}.
\newblock In \emph{Proceedings of the 18th Conference of the European Chapter of the Association for Computational Linguistics (Volume 1: Long Papers)}, pages 226--245, St. Julian{'}s, Malta. Association for Computational Linguistics.

\bibitem[{Ahia et~al.(2023)Ahia, Kumar, Gonen, Kasai, Mortensen, Smith, and Tsvetkov}]{ahia-etal-2023-languages}
Orevaoghene Ahia, Sachin Kumar, Hila Gonen, Jungo Kasai, David Mortensen, Noah Smith, and Yulia Tsvetkov. 2023.
\newblock \href {https://doi.org/10.18653/v1/2023.emnlp-main.614} {Do all languages cost the same? tokenization in the era of commercial language models}.
\newblock In \emph{Proceedings of the 2023 Conference on Empirical Methods in Natural Language Processing}, pages 9904--9923, Singapore. Association for Computational Linguistics.

\bibitem[{Conneau et~al.(2020)Conneau, Khandelwal, Goyal, Chaudhary, Wenzek, Guzm{\'a}n, Grave, Ott, Zettlemoyer, and Stoyanov}]{xlmr}
Alexis Conneau, Kartikay Khandelwal, Naman Goyal, Vishrav Chaudhary, Guillaume Wenzek, Francisco Guzm{\'a}n, Edouard Grave, Myle Ott, Luke Zettlemoyer, and Veselin Stoyanov. 2020.
\newblock \href {https://doi.org/10.18653/v1/2020.acl-main.747} {Unsupervised cross-lingual representation learning at scale}.
\newblock In \emph{Proceedings of the 58th Annual Meeting of the Association for Computational Linguistics}, pages 8440--8451, Online. Association for Computational Linguistics.

\bibitem[{Eberhard et~al.(2025)Eberhard, Simons, and Fennig}]{ethnologue28}
David~M. Eberhard, Gary~F. Simons, and Charles~D. Fennig, editors. 2025.
\newblock \href {https://www.ethnologue.com} {\emph{Ethnologue: Languages of the World}}, 28th edition.
\newblock SIL International, Dallas, TX.

\bibitem[{{Gemma Team}(2025)}]{gemma}
{Gemma Team}. 2025.
\newblock \href {https://goo.gle/Gemma3Report} {Gemma 3}.

\bibitem[{Grattafiori et~al.(2024)Grattafiori, Dubey, Jauhri, Pandey, Kadian, Al-Dahle, Letman, Mathur, Schelten, Vaughan et~al.}]{llama}
Aaron Grattafiori, Abhimanyu Dubey, Abhinav Jauhri, Abhinav Pandey, Abhishek Kadian, Ahmad Al-Dahle, Aiesha Letman, Akhil Mathur, Alan Schelten, Alex Vaughan, and 1 others. 2024.
\newblock The llama 3 herd of models.
\newblock \emph{arXiv preprint arXiv:2407.21783}.

\bibitem[{Imani et~al.(2023)Imani, Lin, Kargaran, Severini, Jalili~Sabet, Kassner, Ma, Schmid, Martins, Yvon, and Sch{\"u}tze}]{imanigooghari-etal-2023-glot500}
Ayyoob Imani, Peiqin Lin, Amir~Hossein Kargaran, Silvia Severini, Masoud Jalili~Sabet, Nora Kassner, Chunlan Ma, Helmut Schmid, Andr{\'e} Martins, Fran{\c{c}}ois Yvon, and Hinrich Sch{\"u}tze. 2023.
\newblock \href {https://doi.org/10.18653/v1/2023.acl-long.61} {Glot500: Scaling multilingual corpora and language models to 500 languages}.
\newblock In \emph{Proceedings of the 61st Annual Meeting of the Association for Computational Linguistics (Volume 1: Long Papers)}, pages 1082--1117, Toronto, Canada. Association for Computational Linguistics.

\bibitem[{Jalili~Sabet et~al.(2020)Jalili~Sabet, Dufter, Yvon, and Sch{\"u}tze}]{simalign}
Masoud Jalili~Sabet, Philipp Dufter, Fran{\c{c}}ois Yvon, and Hinrich Sch{\"u}tze. 2020.
\newblock \href {https://www.aclweb.org/anthology/2020.findings-emnlp.147} {{S}im{A}lign: High quality word alignments without parallel training data using static and contextualized embeddings}.
\newblock In \emph{Proceedings of the 2020 Conference on Empirical Methods in Natural Language Processing: Findings}, pages 1627--1643, Online. Association for Computational Linguistics.

\bibitem[{Martins et~al.(2025)Martins, Alves, Fernandes, Guerreiro, Rei, Farajian, Klimaszewski, Alves, Pombal, Boizard et~al.}]{eurollm}
Pedro~Henrique Martins, Jo{\~a}o Alves, Patrick Fernandes, Nuno~M Guerreiro, Ricardo Rei, Amin Farajian, Mateusz Klimaszewski, Duarte~M Alves, Jos{\'e} Pombal, Nicolas Boizard, and 1 others. 2025.
\newblock Eurollm-9b: Technical report.
\newblock \emph{arXiv preprint arXiv:2506.04079}.

\bibitem[{Nguyen et~al.(2024)Nguyen, Nguyen, Lai, Man, Ngo, Dernoncourt, Rossi, and Nguyen}]{culturax}
Thuat Nguyen, Chien~Van Nguyen, Viet~Dac Lai, Hieu Man, Nghia~Trung Ngo, Franck Dernoncourt, Ryan~A. Rossi, and Thien~Huu Nguyen. 2024.
\newblock \href {https://aclanthology.org/2024.lrec-main.377/} {{C}ultura{X}: A cleaned, enormous, and multilingual dataset for large language models in 167 languages}.
\newblock In \emph{Proceedings of the 2024 Joint International Conference on Computational Linguistics, Language Resources and Evaluation (LREC-COLING 2024)}, pages 4226--4237, Torino, Italia. ELRA and ICCL.

\bibitem[{{NLLB Team} et~al.(2024){NLLB Team}, Costa-juss{\`a}, Cross, {\c{C}}elebi, Elbayad, Heafield, Heffernan, Kalbassi, Lam, Licht, Maillard, Sun, Wang, Wenzek, Youngblood, Akula, Barrault, Gonzalez, Hansanti, Hoffman, Jarrett, Sadagopan, Rowe, Spruit, Tran, Andrews, Ayan, Bhosale, Edunov, Fan, Gao, Goswami, Guzm{\'a}n, Koehn, Mourachko, Ropers, Saleem, Schwenk, and Wang}]{flores}
{NLLB Team}, Marta~R. Costa-juss{\`a}, James Cross, Onur {\c{C}}elebi, Maha Elbayad, Kenneth Heafield, Kevin Heffernan, Elahe Kalbassi, Janice Lam, Daniel Licht, Jean Maillard, Anna Sun, Skyler Wang, Guillaume Wenzek, Al~Youngblood, Bapi Akula, Loic Barrault, Gabriel~Mejia Gonzalez, Prangthip Hansanti, and 20 others. 2024.
\newblock \href {https://doi.org/10.1038/s41586-024-07335-x} {Scaling neural machine translation to 200 languages}.
\newblock \emph{Nature}, 630(8018):841--846.

\bibitem[{Ortiz~Su{'a}rez et~al.(2020)Ortiz~Su{'a}rez, Romary, and Sagot}]{oscar}
Pedro~Javier Ortiz~Su{'a}rez, Laurent Romary, and Benoit Sagot. 2020.
\newblock \href {https://www.aclweb.org/anthology/2020.acl-main.156} {A monolingual approach to contextualized word embeddings for mid-resource languages}.
\newblock In \emph{Proceedings of the 58th Annual Meeting of the Association for Computational Linguistics}, pages 1703--1714, Online. Association for Computational Linguistics.

\bibitem[{{\"O}stling and Tiedemann(2016)}]{eflomal}
Robert {\"O}stling and J{\"o}rg Tiedemann. 2016.
\newblock \href {http://ufal.mff.cuni.cz/pbml/106/art-ostling-tiedemann.pdf} {Efficient word alignment with {M}arkov {C}hain {M}onte {C}arlo}.
\newblock \emph{Prague Bulletin of Mathematical Linguistics}, 106:125--146.

\bibitem[{Penedo et~al.(2025)Penedo, Kydlíček, Sabolčec, Messmer, Foroutan, Kargaran, Raffel, Jaggi, Werra, and Wolf}]{fineweb2}
Guilherme Penedo, Hynek Kydlíček, Vinko Sabolčec, Bettina Messmer, Negar Foroutan, Amir~Hossein Kargaran, Colin Raffel, Martin Jaggi, Leandro~Von Werra, and Thomas Wolf. 2025.
\newblock \href {https://arxiv.org/abs/2506.20920} {Fineweb2: One pipeline to scale them all -- adapting pre-training data processing to every language}.
\newblock \emph{Preprint}, arXiv:2506.20920.

\bibitem[{Petrov et~al.(2023)Petrov, Malfa, Torr, and Bibi}]{parity}
Aleksandar Petrov, Emanuele~La Malfa, Philip~H.S. Torr, and Adel Bibi. 2023.
\newblock Language model tokenizers introduce unfairness between languages.
\newblock In \emph{Proceedings of the 37th International Conference on Neural Information Processing Systems}, NIPS '23, Red Hook, NY, USA. Curran Associates Inc.

\bibitem[{{Qwen Team}(2025)}]{qwen}
{Qwen Team}. 2025.
\newblock \href {https://arxiv.org/abs/2505.09388} {Qwen3 technical report}.
\newblock \emph{Preprint}, arXiv:2505.09388.

\bibitem[{Rabanus et~al.(2023--)Rabanus, Kruijt, Alber, Bidese, Gaeta, Raimondi, Mas, Bertollo, Bissolo, Bonelli, Capelli, Casalicchio, Cioffi, Cordin, Cosentino, {Dal Negro}, Driussi, Glück, Kokkelmans, Murelli, Padovan, Pons, Rivoira, Tagliani, Saracco, Tomaselli, Videsott, Vietti, and Vogt}]{alpilink}
Stefan Rabanus, Anne Kruijt, Birgit Alber, Ermenegildo Bidese, Livio Gaeta, Gianmario Raimondi, Paolo~Benedetto Mas, Sabrina Bertollo, Serena Bissolo, Angelica Bonelli, Dario Capelli, Jan Casalicchio, Raffaele Cioffi, Patrizia Cordin, Michele Cosentino, Silvia {Dal Negro}, Ilaria Driussi, Alexander Glück, Joachim Kokkelmans, and 10 others. 2023--.
\newblock {AlpiLinK. German-Romance language contact in the Italian Alps: documentation, explanation, participation}.
\newblock \url{https://alpilink.it}.
\newblock Ongoing project.

\bibitem[{Raffel et~al.(2020)Raffel, Shazeer, Roberts, Lee, Narang, Matena, Zhou, Li, and Liu}]{c4}
Colin Raffel, Noam Shazeer, Adam Roberts, Katherine Lee, Sharan Narang, Michael Matena, Yanqi Zhou, Wei Li, and Peter~J. Liu. 2020.
\newblock Exploring the limits of transfer learning with a unified text-to-text transformer.
\newblock \emph{J. Mach. Learn. Res.}, 21(1).

\bibitem[{Ramponi and Casula(2023)}]{diatopit}
Alan Ramponi and Camilla Casula. 2023.
\newblock \href {https://doi.org/10.18653/v1/2023.vardial-1.19} {Diatopit: A corpus of social media posts for the study of diatopic language variation in italy}.
\newblock In \emph{Tenth Workshop on NLP for Similar Languages, Varieties and Dialects (VarDial 2023)}, page 187–199, Dubrovnik, Croatia. Association for Computational Linguistics.

\bibitem[{Rei et~al.(2025)Rei, Guerreiro, Pombal, Alves, Teixeirinha, Farajian, and Martins}]{tower}
Ricardo Rei, Nuno~M Guerreiro, Jos{\'e} Pombal, Jo{\~a}o Alves, Pedro Teixeirinha, Amin Farajian, and Andr{\'e}~FT Martins. 2025.
\newblock Tower+: Bridging generality and translation specialization in multilingual llms.
\newblock \emph{arXiv preprint arXiv:2506.17080}.

\bibitem[{Rei et~al.(2020)Rei, Stewart, Farinha, and Lavie}]{rei-etal-2020-comet}
Ricardo Rei, Craig Stewart, Ana~C Farinha, and Alon Lavie. 2020.
\newblock \href {https://doi.org/10.18653/v1/2020.emnlp-main.213} {{COMET}: A neural framework for {MT} evaluation}.
\newblock In \emph{Proceedings of the 2020 Conference on Empirical Methods in Natural Language Processing (EMNLP)}, pages 2685--2702, Online. Association for Computational Linguistics.

\bibitem[{Tiedemann(2020)}]{tatoeba}
J{\"o}rg Tiedemann. 2020.
\newblock \href {https://aclanthology.org/2020.wmt-1.139/} {The tatoeba translation challenge {--} realistic data sets for low resource and multilingual {MT}}.
\newblock In \emph{Proceedings of the Fifth Conference on Machine Translation}, pages 1174--1182, Online. Association for Computational Linguistics.

\bibitem[{{Wikimedia Foundation}()}]{wikidump}
{Wikimedia Foundation}.
\newblock \href {https://dumps.wikimedia.org} {Wikimedia downloads}.

\end{thebibliography}

\appendix

\section{Prompts}\label{app:promps}

\paragraph{Topic classification.}
The system prompt is \texttt{"You are a helpful assistant that classifies the following sentence into one of the following categories: science/technology, travel, politics, sports, health, entertainment, geography. Do not add any explanations."}

The user prompt is \texttt{"Is this a piece of news regarding {"science, technology, travel, politics, sports, health, entertainment, or geography"}? TEXT."}, where \texttt{TEXT} is the sentence we are classifying. For \tower we did not use the system prompt.

\paragraph{Machine translation.}
The system prompt is \texttt{You are a helpful assistant that translates the following sentence from SRG to TGT. Do not add any explanations.}

The user prompt is \texttt{Translate the following SRC source text to TGT:\textbackslash nSRC: SENTENCE\textbackslash nTGT: "}.
\texttt{SRG} and \texttt{TGT} are the name of the source and target language. \texttt{SENTENCE} is the sentence to translate.
For the pivot experiments, the first step translates to Italian, while the second translates from Italian.

\section{Classification Results}\label{app:class}

\Cref{tab:app_class_f1,tab:app_class_prec,tab:app_class_rec,tab:app_class_acc} show the scores on the text classification task.
\begin{table}[H]
    \centering\small
    \begin{tabular}{lcccc}
    \toprule
    \multicolumn{5}{c}{Metric: \fscore } \\
    Model & Pms & Ita & Fra & Eng \\
    \midrule
    \multirow{2}{*}{Llama} & 0.778 & 0.849 & 0.822 & 0.849 \\
     & (0.047) & (0.041) & (0.044) & (0.040) \\
    \multirow{2}{*}{Qwen} & 0.671 & 0.772 & 0.778 & 0.750 \\
     & (0.054) & (0.047) & (0.046) & (0.051) \\
    \multirow{2}{*}{Gemma} & 0.767 & 0.849 & 0.877 & 0.863 \\
     & (0.050) & (0.042) & (0.037) & (0.041) \\
    \multirow{2}{*}{Tower} & 0.589 & 0.781 & 0.781 & 0.795 \\
     & (0.057) & (0.049) & (0.049) & (0.046) \\
    \multirow{2}{*}{Eurollm} & 0.579 & 0.524 & 0.479 & 0.671 \\
     & (0.058) & (0.058) & (0.057) & (0.054) \\
    \multirow{2}{*}{Gemini} & 0.822 & 0.904 & 0.877 & 0.918 \\
     & (0.044) & (0.034) & (0.037) & (0.031) \\
    \multirow{2}{*}{Gpt} & 0.772 & 0.808 & 0.822 & 0.836 \\
     & (0.048) & (0.045) & (0.045) & (0.042) \\
    \bottomrule
    \end{tabular}
    \caption{\fscore of the different models on the classification task. In parenthesis the STD of the score.}
    \label{tab:app_class_f1}
\end{table}

\begin{table}[H]
    \centering\small
\begin{tabular}{lcccc}
\toprule
\multicolumn{5}{c}{Metric: Precision } \\
Model & Pms & Ita & Fra & Eng \\
\midrule
\multirow{2}{*}{Llama} & 0.789 & 0.849 & 0.822 & 0.849 \\
 & (0.047) & (0.041) & (0.044) & (0.040) \\
\multirow{2}{*}{Qwen} & 0.686 & 0.778 & 0.789 & 0.761 \\
 & (0.054) & (0.047) & (0.046) & (0.051) \\
\multirow{2}{*}{Gemma} & 0.767 & 0.849 & 0.877 & 0.863 \\
 & (0.050) & (0.042) & (0.037) & (0.041) \\
\multirow{2}{*}{Tower} & 0.589 & 0.781 & 0.781 & 0.795 \\
 & (0.057) & (0.049) & (0.049) & (0.046) \\
\multirow{2}{*}{Eurollm} & 0.583 & 0.528 & 0.479 & 0.671 \\
 & (0.058) & (0.058) & (0.057) & (0.054) \\
\multirow{2}{*}{Gemini} & 0.822 & 0.904 & 0.877 & 0.918 \\
 & (0.044) & (0.034) & (0.037) & (0.031) \\
\multirow{2}{*}{Gpt} & 0.778 & 0.808 & 0.822 & 0.836 \\
 & (0.048) & (0.045) & (0.045) & (0.042) \\
\bottomrule
\end{tabular}
    \caption{Precision of the different models on the classification task. In parenthesis the STD of the score.}
    \label{tab:app_class_prec}
\end{table}

\begin{table}[H]
    \centering\small
\begin{tabular}{lcccc}
\toprule
\multicolumn{5}{c}{Metric: Recall } \\
Model & Pms & Ita & Fra & Eng \\
\midrule
\multirow{2}{*}{Llama} & 0.767 & 0.849 & 0.822 & 0.849 \\
 & (0.048) & (0.041) & (0.044) & (0.040) \\
\multirow{2}{*}{Qwen} & 0.658 & 0.767 & 0.767 & 0.740 \\
 & (0.054) & (0.047) & (0.047) & (0.052) \\
\multirow{2}{*}{Gemma} & 0.767 & 0.849 & 0.877 & 0.863 \\
 & (0.050) & (0.042) & (0.037) & (0.041) \\
\multirow{2}{*}{Tower} & 0.589 & 0.781 & 0.781 & 0.795 \\
 & (0.057) & (0.049) & (0.049) & (0.046) \\
\multirow{2}{*}{Eurollm} & 0.575 & 0.521 & 0.479 & 0.671 \\
 & (0.058) & (0.058) & (0.057) & (0.054) \\
\multirow{2}{*}{Gemini} & 0.822 & 0.904 & 0.877 & 0.918 \\
 & (0.044) & (0.034) & (0.037) & (0.031) \\
\multirow{2}{*}{Gpt} & 0.767 & 0.808 & 0.822 & 0.836 \\
 & (0.048) & (0.045) & (0.045) & (0.042) \\
\bottomrule
\end{tabular}
    \caption{Recall of the different models on the classification task. In parenthesis the STD of the score.}
    \label{tab:app_class_rec}
\end{table}

\begin{table}[H]
    \centering\small
\begin{tabular}{lcccc}
\toprule
\multicolumn{5}{c}{Metric: Accuracy } \\
Model & Pms & Ita & Fra & Eng \\
\midrule
\multirow{2}{*}{Llama} & 0.767 & 0.849 & 0.822 & 0.849 \\
 & (0.048) & (0.041) & (0.044) & (0.040) \\
\multirow{2}{*}{Qwen} & 0.658 & 0.767 & 0.767 & 0.740 \\
 & (0.054) & (0.047) & (0.047) & (0.052) \\
\multirow{2}{*}{Gemma} & 0.767 & 0.849 & 0.877 & 0.863 \\
 & (0.050) & (0.042) & (0.037) & (0.041) \\
\multirow{2}{*}{Tower} & 0.589 & 0.781 & 0.781 & 0.795 \\
 & (0.057) & (0.049) & (0.049) & (0.046) \\
\multirow{2}{*}{Eurollm} & 0.575 & 0.521 & 0.479 & 0.671 \\
 & (0.058) & (0.058) & (0.057) & (0.054) \\
\multirow{2}{*}{Gemini} & 0.822 & 0.904 & 0.877 & 0.918 \\
 & (0.044) & (0.034) & (0.037) & (0.031) \\
\multirow{2}{*}{Gpt} & 0.767 & 0.808 & 0.822 & 0.836 \\
 & (0.048) & (0.045) & (0.045) & (0.042) \\
\bottomrule
\end{tabular}
    \caption{Accuracy of the different models on the classification task. In parenthesis the STD of the score.}
    \label{tab:app_class_acc}
\end{table}

\section{Machine Translation Results}\label{app:mt}

\Cref{tab:app_pms_ita,tab:app_pms_fra,tab:app_pms_eng,tab:app_ita_pms,tab:app_ita_fra,tab:app_ita_eng,tab:app_fra_pms,tab:app_fra_ita,tab:app_fra_eng,tab:app_eng_pms,tab:app_eng_ita,tab:app_eng_fra}
show the scores for the direct MT task (standard deviation in parenthesis), while \Cref{tab:app_pms_fra_pivot,tab:app_pms_eng_pivot,tab:app_fra_pms_pivot,tab:app_fra_eng_pivot,tab:app_eng_pms_pivot,tab:app_eng_fra_pivot} show the scores with pivoting.

\begin{table}[H]
\centering\small
\begin{tabular}{lcccc}
\toprule
\multicolumn{5}{c}{Pms → Ita} \\
Model & BLEU & chrF++ & TER & COMET \\
\midrule
\multirow{2}{*}{Llama} & 42.910 & 63.800 & 44.080 & 0.933 \\
 & (3.720) & (2.890) & (3.650) & (0.003) \\
\multirow{2}{*}{Qwen} & 37.490 & 59.360 & 48.890 & 0.932 \\
 & (3.730) & (2.850) & (3.810) & (0.003) \\
\multirow{2}{*}{Gemma} & 42.090 & 62.920 & 45.960 & 0.935 \\
 & (4.260) & (3.110) & (4.300) & (0.003) \\
\multirow{2}{*}{Tower} & 34.450 & 57.450 & 56.690 & 0.931 \\
 & (4.100) & (3.090) & (5.850) & (0.003) \\
\multirow{2}{*}{Eurollm} & 32.290 & 56.140 & 56.130 & 0.909 \\
 & (3.690) & (3.110) & (4.300) & (0.009) \\
\multirow{2}{*}{Gemini} & 46.290 & 66.640 & 42.450 & 0.937 \\
 & (4.310) & (3.130) & (5.390) & (0.002) \\
\multirow{2}{*}{Gpt} & 42.290 & 63.040 & 44.300 & 0.932 \\
 & (4.170) & (3.230) & (4.060) & (0.004) \\
\bottomrule
\end{tabular}
\caption{Translation results from Pms to Ita.}
\label{tab:app_pms_ita}
\end{table}

\begin{table}[H]
\centering\small
\begin{tabular}{lcccc}
\toprule
\multicolumn{5}{c}{Pms → Fra} \\
Model & BLEU & chrF++ & TER & COMET \\
\midrule
\multirow{2}{*}{Llama} & 21.700 & 47.070 & 70.330 & 0.908 \\
 & (2.400) & (2.160) & (3.470) & (0.005) \\
\multirow{2}{*}{Qwen} & 20.420 & 45.510 & 71.000 & 0.905 \\
 & (2.500) & (2.160) & (2.920) & (0.005) \\
\multirow{2}{*}{Gemma} & 21.550 & 47.170 & 70.680 & 0.908 \\
 & (2.640) & (2.270) & (3.490) & (0.005) \\
\multirow{2}{*}{Tower} & 17.770 & 43.320 & 77.660 & 0.906 \\
 & (2.310) & (2.280) & (3.690) & (0.005) \\
\multirow{2}{*}{Eurollm} & 16.670 & 41.550 & 79.670 & 0.901 \\
 & (2.500) & (2.630) & (3.580) & (0.006) \\
\multirow{2}{*}{Gemini} & 21.550 & 47.850 & 71.460 & 0.908 \\
 & (2.690) & (2.160) & (4.840) & (0.005) \\
\multirow{2}{*}{Gpt} & 20.670 & 45.980 & 69.420 & 0.906 \\
 & (2.410) & (2.200) & (3.110) & (0.005) \\
\bottomrule
\end{tabular}
\caption{Translation results from Pms to Fra.}
\label{tab:app_pms_fra}
\end{table}

\begin{table}[H]
\centering\small
\begin{tabular}{lcccc}
\toprule
\multicolumn{5}{c}{Pms → Eng} \\
Model & BLEU & chrF++ & TER & COMET \\
\midrule
\multirow{2}{*}{Llama} & 20.270 & 48.640 & 71.780 & 0.901 \\
 & (2.760) & (2.490) & (4.460) & (0.004) \\
\multirow{2}{*}{Qwen} & 19.180 & 46.060 & 74.230 & 0.900 \\
 & (2.690) & (2.380) & (3.790) & (0.004) \\
\multirow{2}{*}{Gemma} & 19.970 & 48.200 & 73.510 & 0.900 \\
 & (2.810) & (2.550) & (4.220) & (0.004) \\
\multirow{2}{*}{Tower} & 17.540 & 45.230 & 80.070 & 0.900 \\
 & (2.620) & (2.570) & (4.900) & (0.004) \\
\multirow{2}{*}{Eurollm} & 16.080 & 44.180 & 82.440 & 0.894 \\
 & (2.750) & (2.420) & (4.930) & (0.006) \\
\multirow{2}{*}{Gemini} & 22.800 & 51.240 & 68.290 & 0.899 \\
 & (2.680) & (2.280) & (3.800) & (0.004) \\
\multirow{2}{*}{Gpt} & 20.940 & 48.770 & 71.950 & 0.899 \\
 & (2.720) & (2.380) & (3.760) & (0.005) \\
\bottomrule
\end{tabular}
\caption{Translation results from Pms to Eng.}
\label{tab:app_pms_eng}
\end{table}

\begin{table}[H]
\centering\small
\begin{tabular}{lcccc}
\toprule
\multicolumn{5}{c}{Ita → Pms} \\
Model & BLEU & chrF++ & TER & COMET \\
\midrule
\multirow{2}{*}{Llama} & 6.980 & 34.780 & 83.220 & 0.842 \\
 & (1.290) & (1.640) & (2.550) & (0.008) \\
\multirow{2}{*}{Qwen} & 5.250 & 31.100 & 93.290 & 0.843 \\
 & (1.540) & (1.760) & (7.820) & (0.007) \\
\multirow{2}{*}{Gemma} & 6.270 & 34.090 & 84.580 & 0.841 \\
 & (1.310) & (1.560) & (2.560) & (0.008) \\
\multirow{2}{*}{Tower} & 4.560 & 31.200 & 99.970 & 0.843 \\
 & (1.240) & (1.850) & (12.440) & (0.007) \\
\multirow{2}{*}{Eurollm} & 6.250 & 33.870 & 88.590 & 0.844 \\
 & (1.830) & (2.020) & (12.490) & (0.007) \\
\multirow{2}{*}{Gemini} & 7.280 & 33.930 & 85.080 & 0.833 \\
 & (1.470) & (1.820) & (3.030) & (0.009) \\
\multirow{2}{*}{Gpt} & 6.800 & 35.830 & 82.320 & 0.848 \\
 & (1.370) & (1.590) & (2.530) & (0.006) \\
\bottomrule
\end{tabular}
\caption{Translation results from Ita to Pms.}
\label{tab:app_ita_pms}
\end{table}

\begin{table}[H]
\centering\small
\begin{tabular}{lcccc}
\toprule
\multicolumn{5}{c}{Ita → Fra} \\
Model & BLEU & chrF++ & TER & COMET \\
\midrule
\multirow{2}{*}{Llama} & 33.500 & 58.890 & 54.750 & 0.943 \\
 & (2.550) & (1.840) & (2.930) & (0.003) \\
\multirow{2}{*}{Qwen} & 32.380 & 58.060 & 55.300 & 0.943 \\
 & (2.510) & (1.750) & (2.940) & (0.003) \\
\multirow{2}{*}{Gemma} & 34.260 & 59.610 & 53.520 & 0.942 \\
 & (2.720) & (1.890) & (3.150) & (0.003) \\
\multirow{2}{*}{Tower} & 34.100 & 59.110 & 53.900 & 0.943 \\
 & (2.660) & (1.900) & (2.900) & (0.003) \\
\multirow{2}{*}{Eurollm} & 34.700 & 60.190 & 53.000 & 0.944 \\
 & (2.450) & (1.810) & (2.990) & (0.003) \\
\multirow{2}{*}{Gemini} & 33.100 & 58.830 & 55.880 & 0.943 \\
 & (2.710) & (1.750) & (3.090) & (0.003) \\
\multirow{2}{*}{Gpt} & 32.400 & 58.600 & 55.270 & 0.942 \\
 & (2.320) & (1.700) & (2.710) & (0.003) \\
\bottomrule
\end{tabular}
\caption{Translation results from Ita to Fra.}
\label{tab:app_ita_fra}
\end{table}

\begin{table}[H]
\centering\small
\begin{tabular}{lcccc}
\toprule
\multicolumn{5}{c}{Ita → Eng} \\
Model & BLEU & chrF++ & TER & COMET \\
\midrule
\multirow{2}{*}{Llama} & 32.520 & 60.930 & 52.970 & 0.941 \\
 & (2.910) & (2.050) & (3.430) & (0.003) \\
\multirow{2}{*}{Qwen} & 32.050 & 60.750 & 52.360 & 0.942 \\
 & (2.790) & (1.920) & (3.110) & (0.003) \\
\multirow{2}{*}{Gemma} & 33.410 & 61.440 & 51.580 & 0.941 \\
 & (2.860) & (1.950) & (3.100) & (0.003) \\
\multirow{2}{*}{Tower} & 33.900 & 61.890 & 51.610 & 0.942 \\
 & (2.800) & (2.020) & (3.160) & (0.003) \\
\multirow{2}{*}{Eurollm} & 32.970 & 60.850 & 52.460 & 0.941 \\
 & (2.810) & (1.920) & (3.140) & (0.003) \\
\multirow{2}{*}{Gemini} & 32.400 & 61.170 & 53.240 & 0.941 \\
 & (2.670) & (1.850) & (3.350) & (0.003) \\
\multirow{2}{*}{Gpt} & 32.730 & 61.390 & 52.430 & 0.941 \\
 & (2.620) & (1.900) & (3.090) & (0.003) \\
\bottomrule
\end{tabular}
\caption{Translation results from Ita to Eng.}
\label{tab:app_ita_eng}
\end{table}

\begin{table}[H]
\centering\small
\begin{tabular}{lcccc}
\toprule
\multicolumn{5}{c}{Fra → Pms} \\
Model & BLEU & chrF++ & TER & COMET \\
\midrule
\multirow{2}{*}{Llama} & 3.000 & 27.160 & 95.790 & 0.835 \\
 & (1.120) & (1.360) & (2.510) & (0.008) \\
\multirow{2}{*}{Qwen} & 2.410 & 24.890 & 101.020 & 0.838 \\
 & (1.110) & (1.400) & (6.610) & (0.007) \\
\multirow{2}{*}{Gemma} & 2.960 & 27.450 & 94.370 & 0.831 \\
 & (0.860) & (1.290) & (2.300) & (0.009) \\
\multirow{2}{*}{Tower} & 2.930 & 27.080 & 97.820 & 0.836 \\
 & (1.120) & (1.530) & (7.040) & (0.008) \\
\multirow{2}{*}{Eurollm} & 2.810 & 26.590 & 95.590 & 0.836 \\
 & (1.100) & (1.370) & (4.060) & (0.008) \\
\multirow{2}{*}{Gemini} & 3.350 & 26.840 & 96.200 & 0.829 \\
 & (0.990) & (1.360) & (2.670) & (0.009) \\
\multirow{2}{*}{Gpt} & 3.280 & 27.080 & 95.270 & 0.835 \\
 & (1.000) & (1.290) & (2.170) & (0.008) \\
\bottomrule
\end{tabular}
\caption{Translation results from Fra to Pms.}
\label{tab:app_fra_pms}
\end{table}

\begin{table}[H]
\centering\small
\begin{tabular}{lcccc}
\toprule
\multicolumn{5}{c}{Fra → Ita} \\
Model & BLEU & chrF++ & TER & COMET \\
\midrule
\multirow{2}{*}{Llama} & 27.890 & 55.010 & 58.320 & 0.952 \\
 & (2.270) & (1.660) & (2.660) & (0.003) \\
\multirow{2}{*}{Qwen} & 26.810 & 54.430 & 59.620 & 0.952 \\
 & (2.450) & (1.750) & (2.710) & (0.003) \\
\multirow{2}{*}{Gemma} & 29.270 & 55.710 & 57.120 & 0.952 \\
 & (2.530) & (1.770) & (2.650) & (0.003) \\
\multirow{2}{*}{Tower} & 29.680 & 55.890 & 57.400 & 0.952 \\
 & (2.690) & (1.910) & (2.800) & (0.003) \\
\multirow{2}{*}{Eurollm} & 29.200 & 55.650 & 57.270 & 0.952 \\
 & (2.500) & (1.660) & (2.670) & (0.003) \\
\multirow{2}{*}{Gemini} & 29.520 & 55.840 & 57.980 & 0.952 \\
 & (2.650) & (1.790) & (3.020) & (0.003) \\
\multirow{2}{*}{Gpt} & 27.680 & 55.140 & 57.640 & 0.951 \\
 & (2.390) & (1.740) & (2.700) & (0.003) \\
\bottomrule
\end{tabular}
\caption{Translation results from Fra to Ita.}
\label{tab:app_fra_ita}
\end{table}

\begin{table}[H]
\centering\small
\begin{tabular}{lcccc}
\toprule
\multicolumn{5}{c}{Fra → Eng} \\
Model & BLEU & chrF++ & TER & COMET \\
\midrule
\multirow{2}{*}{Llama} & 46.510 & 69.120 & 37.150 & 0.947 \\
 & (3.530) & (2.250) & (3.270) & (0.002) \\
\multirow{2}{*}{Qwen} & 45.320 & 68.170 & 38.340 & 0.947 \\
 & (3.510) & (2.130) & (3.400) & (0.002) \\
\multirow{2}{*}{Gemma} & 46.410 & 69.290 & 36.160 & 0.947 \\
 & (3.390) & (2.150) & (3.170) & (0.002) \\
\multirow{2}{*}{Tower} & 45.030 & 68.350 & 37.830 & 0.945 \\
 & (3.330) & (2.190) & (3.170) & (0.002) \\
\multirow{2}{*}{Eurollm} & 47.380 & 69.520 & 36.880 & 0.945 \\
 & (3.810) & (2.370) & (3.480) & (0.002) \\
\multirow{2}{*}{Gemini} & 48.790 & 70.670 & 34.530 & 0.947 \\
 & (3.640) & (2.140) & (3.290) & (0.002) \\
\multirow{2}{*}{Gpt} & 46.460 & 69.410 & 36.770 & 0.947 \\
 & (3.550) & (2.150) & (3.100) & (0.002) \\
\bottomrule
\end{tabular}
\caption{Translation results from Fra to Eng.}
\label{tab:app_fra_eng}
\end{table}

\begin{table}[H]
\centering\small
\begin{tabular}{lcccc}
\toprule
\multicolumn{5}{c}{Eng → Pms} \\
Model & BLEU & chrF++ & TER & COMET \\
\midrule
\multirow{2}{*}{Llama} & 3.170 & 28.010 & 94.430 & 0.824 \\
 & (1.140) & (1.390) & (2.370) & (0.009) \\
\multirow{2}{*}{Qwen} & 2.340 & 24.970 & 98.110 & 0.828 \\
 & (1.200) & (1.280) & (3.870) & (0.008) \\
\multirow{2}{*}{Gemma} & 2.460 & 27.470 & 94.370 & 0.825 \\
 & (0.840) & (1.300) & (2.370) & (0.008) \\
\multirow{2}{*}{Tower} & 3.170 & 27.620 & 99.740 & 0.830 \\
 & (1.140) & (1.680) & (9.360) & (0.008) \\
\multirow{2}{*}{Eurollm} & 3.050 & 27.860 & 94.830 & 0.828 \\
 & (1.070) & (1.420) & (6.580) & (0.008) \\
\multirow{2}{*}{Gemini} & 3.130 & 26.980 & 95.730 & 0.818 \\
 & (0.890) & (1.470) & (4.030) & (0.010) \\
\multirow{2}{*}{Gpt} & 2.290 & 27.990 & 92.130 & 0.830 \\
 & (0.920) & (1.220) & (1.940) & (0.007) \\
\bottomrule
\end{tabular}
\caption{Translation results from Eng to Pms.}
\label{tab:app_eng_pms}
\end{table}

\begin{table}[H]
\centering\small
\begin{tabular}{lcccc}
\toprule
\multicolumn{5}{c}{Eng → Ita} \\
Model & BLEU & chrF++ & TER & COMET \\
\midrule
\multirow{2}{*}{Llama} & 29.850 & 56.230 & 54.900 & 0.953 \\
 & (2.110) & (1.650) & (2.390) & (0.003) \\
\multirow{2}{*}{Qwen} & 29.920 & 56.340 & 56.200 & 0.952 \\
 & (2.560) & (1.930) & (2.790) & (0.003) \\
\multirow{2}{*}{Gemma} & 30.770 & 57.210 & 54.380 & 0.953 \\
 & (2.510) & (1.760) & (2.430) & (0.003) \\
\multirow{2}{*}{Tower} & 34.510 & 59.740 & 51.570 & 0.953 \\
 & (3.070) & (1.990) & (2.610) & (0.003) \\
\multirow{2}{*}{Eurollm} & 33.590 & 58.860 & 52.000 & 0.953 \\
 & (2.530) & (1.740) & (2.500) & (0.003) \\
\multirow{2}{*}{Gemini} & 31.660 & 58.100 & 53.300 & 0.953 \\
 & (2.320) & (1.550) & (2.430) & (0.003) \\
\multirow{2}{*}{Gpt} & 31.240 & 57.750 & 52.740 & 0.952 \\
 & (2.340) & (1.620) & (2.320) & (0.003) \\
\bottomrule
\end{tabular}
\caption{Translation results from Eng to Ita.}
\label{tab:app_eng_ita}
\end{table}

\begin{table}[H]
\centering\small
\begin{tabular}{lcccc}
\toprule
\multicolumn{5}{c}{Eng → Fra} \\
Model & BLEU & chrF++ & TER & COMET \\
\midrule
\multirow{2}{*}{Llama} & 52.800 & 71.170 & 33.280 & 0.949 \\
 & (3.640) & (2.420) & (2.970) & (0.002) \\
\multirow{2}{*}{Qwen} & 48.010 & 68.150 & 38.380 & 0.948 \\
 & (3.130) & (1.980) & (2.820) & (0.002) \\
\multirow{2}{*}{Gemma} & 52.420 & 71.090 & 33.750 & 0.947 \\
 & (3.510) & (2.360) & (2.920) & (0.003) \\
\multirow{2}{*}{Tower} & 53.020 & 71.410 & 34.360 & 0.949 \\
 & (3.360) & (2.130) & (2.800) & (0.002) \\
\multirow{2}{*}{Eurollm} & 55.550 & 73.030 & 32.350 & 0.950 \\
 & (3.680) & (2.280) & (3.040) & (0.002) \\
\multirow{2}{*}{Gemini} & 55.030 & 72.550 & 32.150 & 0.945 \\
 & (3.690) & (2.380) & (3.060) & (0.003) \\
\multirow{2}{*}{Gpt} & 53.010 & 71.610 & 33.280 & 0.950 \\
 & (3.260) & (2.110) & (2.890) & (0.002) \\
\bottomrule
\end{tabular}
\caption{Translation results from Eng to Fra.}
\label{tab:app_eng_fra}
\end{table}

\begin{table}[H]
\centering\small
\begin{tabular}{lcccc}
\toprule
\multicolumn{5}{c}{Pms → Fra, Pivot} \\
Model & BLEU & chrF++ & TER & COMET \\
\midrule
\multirow{2}{*}{Llama} & 23.580 & 49.090 & 68.000 & 0.927 \\
 & (2.450) & (2.220) & (3.290) & (0.002) \\
\multirow{2}{*}{Qwen} & 19.890 & 45.140 & 72.630 & 0.927 \\
 & (2.460) & (2.180) & (3.190) & (0.002) \\
\multirow{2}{*}{Gemma} & 23.740 & 49.450 & 68.140 & 0.927 \\
 & (2.820) & (2.290) & (3.570) & (0.002) \\
\multirow{2}{*}{Tower} & 20.380 & 46.070 & 73.850 & 0.927 \\
 & (2.460) & (2.150) & (3.860) & (0.002) \\
\multirow{2}{*}{Eurollm} & 20.210 & 44.900 & 74.080 & 0.927 \\
 & (2.850) & (2.540) & (3.840) & (0.002) \\
\multirow{2}{*}{Gemini} & 25.600 & 50.780 & 64.740 & 0.927 \\
 & (2.410) & (2.150) & (2.930) & (0.002) \\
\multirow{2}{*}{Gpt} & 23.930 & 49.320 & 66.070 & 0.927 \\
 & (2.270) & (2.090) & (2.940) & (0.002) \\
\bottomrule
\end{tabular}
\caption{Translation results with pivoting from Pms to Fra.}
\label{tab:app_pms_fra_pivot}
\end{table}

\begin{table}[H]
\centering\small
\begin{tabular}{lcccc}
\toprule
\multicolumn{5}{c}{Pms → Eng, Pivot} \\
Model & BLEU & chrF++ & TER & COMET \\
\midrule
\multirow{2}{*}{Llama} & 21.780 & 49.550 & 68.830 & 0.917 \\
 & (2.910) & (2.450) & (3.870) & (0.002) \\
\multirow{2}{*}{Qwen} & 19.340 & 47.160 & 70.800 & 0.917 \\
 & (2.730) & (2.400) & (3.610) & (0.002) \\
\multirow{2}{*}{Gemma} & 23.630 & 50.770 & 66.010 & 0.917 \\
 & (3.050) & (2.500) & (4.070) & (0.002) \\
\multirow{2}{*}{Tower} & 18.650 & 47.080 & 75.450 & 0.917 \\
 & (2.740) & (2.430) & (4.590) & (0.002) \\
\multirow{2}{*}{Eurollm} & 18.140 & 45.500 & 77.720 & 0.917 \\
 & (2.940) & (2.570) & (4.310) & (0.002) \\
\multirow{2}{*}{Gemini} & 25.320 & 53.280 & 64.010 & 0.917 \\
 & (2.730) & (2.360) & (3.740) & (0.002) \\
\multirow{2}{*}{Gpt} & 23.490 & 50.760 & 66.860 & 0.917 \\
 & (2.950) & (2.440) & (3.950) & (0.002) \\
\bottomrule
\end{tabular}
\caption{Translation results with pivoting from Pms to Eng.}
\label{tab:app_pms_eng_pivot}
\end{table}

\begin{table}[H]
\centering\small
\begin{tabular}{lcccc}
\toprule
\multicolumn{5}{c}{Fra → Pms, Pivot} \\
Model & BLEU & chrF++ & TER & COMET \\
\midrule
\multirow{2}{*}{Llama} & 3.990 & 28.780 & 92.600 & 0.836 \\
 & (1.180) & (1.430) & (2.440) & (0.008) \\
\multirow{2}{*}{Qwen} & 2.720 & 26.220 & 99.710 & 0.838 \\
 & (1.120) & (1.430) & (7.940) & (0.007) \\
\multirow{2}{*}{Gemma} & 3.760 & 28.640 & 92.540 & 0.834 \\
 & (1.060) & (1.430) & (2.410) & (0.008) \\
\multirow{2}{*}{Tower} & 2.460 & 26.440 & 107.140 & 0.837 \\
 & (1.040) & (1.590) & (15.430) & (0.008) \\
\multirow{2}{*}{Eurollm} & 2.870 & 26.960 & 95.120 & 0.835 \\
 & (1.060) & (1.480) & (5.920) & (0.008) \\
\multirow{2}{*}{Gemini} & 3.880 & 28.000 & 94.160 & 0.829 \\
 & (1.110) & (1.550) & (2.800) & (0.009) \\
\multirow{2}{*}{Gpt} & 3.580 & 29.110 & 93.090 & 0.838 \\
 & (1.080) & (1.400) & (2.280) & (0.007) \\
\bottomrule
\end{tabular}
\caption{Translation results with pivoting from Fra to Pms.}
\label{tab:app_fra_pms_pivot}
\end{table}

\begin{table}[H]
\centering\small
\begin{tabular}{lcccc}
\toprule
\multicolumn{5}{c}{Fra → Eng, Pivot} \\
Model & BLEU & chrF++ & TER & COMET \\
\midrule
\multirow{2}{*}{Llama} & 42.730 & 66.800 & 39.970 & 0.952 \\
 & (3.390) & (2.190) & (3.580) & (0.001) \\
\multirow{2}{*}{Qwen} & 42.940 & 65.990 & 40.340 & 0.952 \\
 & (3.490) & (2.240) & (3.410) & (0.001) \\
\multirow{2}{*}{Gemma} & 42.950 & 66.630 & 39.080 & 0.952 \\
 & (3.260) & (2.020) & (3.130) & (0.001) \\
\multirow{2}{*}{Tower} & 40.260 & 64.990 & 42.240 & 0.952 \\
 & (3.210) & (2.100) & (3.450) & (0.001) \\
\multirow{2}{*}{Eurollm} & 41.720 & 65.570 & 42.000 & 0.952 \\
 & (3.450) & (2.250) & (3.750) & (0.001) \\
\multirow{2}{*}{Gemini} & 43.330 & 67.260 & 39.590 & 0.952 \\
 & (3.650) & (2.150) & (3.400) & (0.001) \\
\multirow{2}{*}{Gpt} & 42.820 & 66.820 & 39.970 & 0.952 \\
 & (3.490) & (2.320) & (3.490) & (0.001) \\
\bottomrule
\end{tabular}
\caption{Translation results with pivoting from Fra to Eng.}
\label{tab:app_fra_eng_pivot}
\end{table}

\begin{table}[H]
\centering\small
\begin{tabular}{lcccc}
\toprule
\multicolumn{5}{c}{Eng → Pms, Pivot} \\
Model & BLEU & chrF++ & TER & COMET \\
\midrule
\multirow{2}{*}{Llama} & 3.770 & 28.910 & 92.680 & 0.826 \\
 & (1.050) & (1.410) & (2.490) & (0.008) \\
\multirow{2}{*}{Qwen} & 2.200 & 26.200 & 96.920 & 0.828 \\
 & (0.980) & (1.370) & (3.500) & (0.008) \\
\multirow{2}{*}{Gemma} & 3.560 & 29.210 & 91.640 & 0.826 \\
 & (1.040) & (1.420) & (2.340) & (0.008) \\
\multirow{2}{*}{Tower} & 2.850 & 27.250 & 104.880 & 0.828 \\
 & (1.110) & (1.710) & (11.890) & (0.008) \\
\multirow{2}{*}{Eurollm} & 3.250 & 28.190 & 91.060 & 0.827 \\
 & (1.180) & (1.410) & (1.830) & (0.008) \\
\multirow{2}{*}{Gemini} & 3.970 & 28.850 & 92.650 & 0.822 \\
 & (1.170) & (1.570) & (2.720) & (0.009) \\
\multirow{2}{*}{Gpt} & 3.650 & 29.600 & 90.910 & 0.828 \\
 & (1.100) & (1.380) & (2.120) & (0.008) \\
\bottomrule
\end{tabular}
\caption{Translation results with pivoting from Eng to Pms.}
\label{tab:app_eng_pms_pivot}
\end{table}

\begin{table}[H]
\centering\small
\begin{tabular}{lcccc}
\toprule
\multicolumn{5}{c}{Eng → Fra, Pivot} \\
Model & BLEU & chrF++ & TER & COMET \\
\midrule
\multirow{2}{*}{Llama} & 47.630 & 67.830 & 37.390 & 0.956 \\
 & (3.130) & (2.170) & (2.810) & (0.001) \\
\multirow{2}{*}{Qwen} & 42.230 & 64.330 & 42.110 & 0.956 \\
 & (2.970) & (1.970) & (2.750) & (0.001) \\
\multirow{2}{*}{Gemma} & 47.460 & 67.640 & 38.090 & 0.956 \\
 & (3.390) & (2.240) & (2.950) & (0.001) \\
\multirow{2}{*}{Tower} & 47.240 & 67.840 & 38.470 & 0.956 \\
 & (3.190) & (2.090) & (2.730) & (0.001) \\
\multirow{2}{*}{Eurollm} & 48.710 & 68.500 & 37.040 & 0.956 \\
 & (3.400) & (2.310) & (3.090) & (0.001) \\
\multirow{2}{*}{Gemini} & 50.750 & 69.800 & 35.350 & 0.956 \\
 & (3.590) & (2.440) & (3.080) & (0.001) \\
\multirow{2}{*}{Gpt} & 48.130 & 68.090 & 37.010 & 0.956 \\
 & (3.350) & (2.330) & (3.030) & (0.001) \\
\bottomrule
\end{tabular}
\caption{Translation results with pivoting from Eng to Fra.}
\label{tab:app_eng_fra_pivot}
\end{table}

\section{Parity Results}\label{app:parity}

\Cref{tab:app_parity_pms,tab:app_parity_ita,tab:app_parity_fra,tab:app_parity_eng} show the tokenizer parity scores with respect to the different languages. Note that scores with respect to different languages are not comparable.

\begin{table}[H]
\centering\small
\begin{tabular}{lccc}
\toprule
\multicolumn{4}{c}{Parity w.r.t Pms} \\
Model & Ita & Fra & Eng \\
\midrule
Llama & 0.905 & 0.898 & 0.569 \\
Qwen & 0.909 & 0.897 & 0.576 \\
Gemma & 0.816 & 0.850 & 0.609 \\
Tower & 0.806 & 0.830 & 0.616 \\
Eurollm & 0.784 & 0.858 & 0.639 \\
Bpe & 0.783 & 0.847 & 0.687 \\
Unigram & 0.752 & 0.817 & 0.664 \\
Gemini & 0.820 & 0.853 & 0.618 \\
Gpt & 0.883 & 0.833 & 0.618 \\
\bottomrule
\end{tabular}
\caption{Parity scores with respect to Piedmontese.}
\label{tab:app_parity_pms}
\end{table}

\begin{table}[H]
\centering\small
\begin{tabular}{lccc}
\toprule
\multicolumn{4}{c}{Parity w.r.t Ita} \\
Model & Pms & Fra & Eng \\
\midrule
Llama & 1.140 & 0.998 & 0.634 \\
Qwen & 1.136 & 0.993 & 0.638 \\
Gemma & 1.267 & 1.041 & 0.749 \\
Tower & 1.284 & 1.033 & 0.767 \\
Eurollm & 1.324 & 1.097 & 0.818 \\
Bpe & 1.320 & 1.088 & 0.881 \\
Unigram & 1.373 & 1.090 & 0.887 \\
Gemini & 1.258 & 1.040 & 0.757 \\
Gpt & 1.162 & 0.945 & 0.703 \\
\bottomrule
\end{tabular}
\caption{Parity scores with respect to Italian.}
\label{tab:app_parity_ita}
\end{table}

\begin{table}[H]
\centering\small
\begin{tabular}{lccc}
\toprule
\multicolumn{4}{c}{Parity w.r.t Fra} \\
Model & Pms & Ita & Eng \\
\midrule
Llama & 1.164 & 1.028 & 0.640 \\
Qwen & 1.167 & 1.034 & 0.649 \\
Gemma & 1.243 & 0.983 & 0.726 \\
Tower & 1.269 & 0.993 & 0.751 \\
Eurollm & 1.232 & 0.935 & 0.753 \\
Bpe & 1.235 & 0.941 & 0.816 \\
Unigram & 1.286 & 0.941 & 0.823 \\
Gemini & 1.235 & 0.984 & 0.734 \\
Gpt & 1.258 & 1.086 & 0.751 \\
\bottomrule
\end{tabular}
\caption{Parity scores with respect to French.}
\label{tab:app_parity_fra}
\end{table}

\begin{table}[H]
\centering\small
\begin{tabular}{lccc}
\toprule
\multicolumn{4}{c}{Parity w.r.t Eng} \\
Model & Pms & Ita & Fra \\
\midrule
Llama & 1.848 & 1.633 & 1.602 \\
Qwen & 1.831 & 1.622 & 1.584 \\
Gemma & 1.740 & 1.380 & 1.416 \\
Tower & 1.721 & 1.347 & 1.371 \\
Eurollm & 1.653 & 1.255 & 1.357 \\
Bpe & 1.532 & 1.166 & 1.250 \\
Unigram & 1.581 & 1.158 & 1.245 \\
Gemini & 1.707 & 1.363 & 1.398 \\
Gpt & 1.699 & 1.470 & 1.366 \\
\bottomrule
\end{tabular}
\caption{Parity scores with respect to English.}
\label{tab:app_parity_eng}
\end{table}

\section{Alignment Metrics}\label{app:align-metric}
We do not use possible reference alignments, so $\lvert P \rvert = \lvert S \rvert$.
Assuming that $\lvert A \cap S \rvert$, $\lvert A \rvert$, and $\lvert S \rvert$ are not empty, $\text{F}_1$ can be rewritten as: 
\begin{align*}
    \text{F}_1 &= \dfrac{2 \text{prec}\cdot\text{rec}}{\text{prec}+\text{rec}} =
    \dfrac{2 \dfrac{\lvert A \cap P \rvert}{\lvert A \rvert}\cdot\dfrac{\lvert A \cap S \rvert}{\lvert S \rvert}}{\dfrac{\lvert A \cap P \rvert}{\lvert A \rvert}+\dfrac{\lvert A \cap S \rvert}{\lvert S \rvert}} = \\
    &= \dfrac{2 \dfrac{\lvert A \cap S \rvert}{\lvert A \rvert}\cdot\dfrac{\lvert A \cap S \rvert}{\lvert S \rvert}}{\dfrac{\lvert A \cap S \rvert}{\lvert A \rvert}+\dfrac{\lvert A \cap S \rvert}{\lvert S \rvert}} = \\
    &= \dfrac{2\lvert A \cap S \rvert}{\lvert A \rvert \lvert S \rvert} \cdot \dfrac{\lvert A \rvert \vert S \rvert}{\lvert A \rvert + \vert S \rvert} =
    \dfrac{2\lvert A \cap S\rvert}{\lvert S\rvert + \lvert A\rvert}
\end{align*}

\noindent And AER as:
\begin{align*}
    \text{AER} = 1 - \dfrac{\lvert A \cap S \rvert + \lvert A \cap P \vert}{\lvert A \rvert + \lvert S \rvert} = 1 - \dfrac{2\lvert A \cap S\rvert}{\lvert S\rvert + \lvert A\rvert}
\end{align*}

Therefore $\text{AER} = 1 - \text{F}_1$


\section{Hyper-parameters} \label{app:hyp}
For the translation task and the classification, we use greedy decoding and we generate at most 100 tokens, which is sufficient for the ground-truth labels. The closed source models do not use reasoning.
The open-weight model are used with the Transformers 4.57.1 text generation pipeline, while the closed models are used through OpenRouter.

We run eflomal (version 2.0.0) with its default parameters, then we symmetrize the alignment with fast align atools with \textit{grow-diag-final-and}.
SimAlign is run with the following arguments:
\begin{itemize}
    \item Model: xlm-roberta-base
    \item Tokenizer type: bpe
    \item Distortion: 0
    \item Layer: 8
    \item Matching method: itermax
\end{itemize}

\section{Computational Resources and Costs}\label{app:cost}
The total cost for Gemini was \$0.62 and \$0.17 for GPT. The provider has a zero data retention policy. The other experiments were run on a local cluster with up to 2 NVIDIA H100 with 95GB VRAM and 60GB RAM.

\section{Questionnaire}\label{app:question}
The questionnaire is in Italian. Here we show the original version and in brackets the English translation. 
The questionnaire introduction explains to the user the goal of the project and emphasizes that it is not about evaluating the user and that it is anonymous.

\onecolumn
\begin{tcolorbox}[colback=Green!20!white,colframe=Green!100!black,breakable,use color stack, title=\textsc{Informazioni su di te e sul piemontese\\\textit{[About you and Piedmontese]}}]




\textbf{Quale lingua utilizzi di più quotidianamente (scuola, lavoro, in giro, ecc.)?}\\ 
\textit{[Which language do you use most on a daily basis?]}\\
Seleziona una lingua dall'elenco. Se scegli "Altro", specifica la lingua nel campo di testo.\\
\textit{[Select a language from the list. If you choose "Other," specify the language in the text field.]}\\ 
\{Seleziona una lingua: Italiano, Francese, Spagnolo,Tedesco, Inglese, Rumeno, Arabo, Macedone, Albanese, Piemontese, Preferisco non rispondere, Altro\}\\
\textit{\{Select a language: Italian, French, Spanish, German, English, Romanian, Arabic, Macedonian, Albanian, Piedmontese, Prefer not to reply, Other\}}

\textbf{Quanto bene parli il piemontese?}\\
\textit{[How well do you speak Peidmontese]}\\
Seleziona una delle opzioni che meglio descrive la tua conoscenza del piemontese.\\
\textit{[Select one of the options that best describes your knowledge of Piedmontese.]}
\begin{compactitem}
    \item Niente o quasi, solo qualche parola \textit{[Nothing or almost nothing, just a few words]}
    \item Poco, conosco alcune espressioni, ma faccio fatica a esprimere frasi nuove \textit{[Not much, I know some expressions, but I struggle to express new sentences]}
    \item Abbastanza, ma a volte lo mischio con l'italiano (o la lingua che uso principalmente) \textit{[Quite a bit, but sometimes I mix it with Italian (or whatever language I use mostly)]}
    \item Perfettamente o quasi, riesco a esprimere praticamente tutto \textit{[Perfectly or almost perfectly, I can express practically everything]}
\end{compactitem}

\textbf{Secondo te, il piemontese ha una grammatica e ortografia ben definita ("questa parola si scrive così", "questo verbo si coniuga cosà")?}\\
\textit{[In your opinion, does Piedmontese have a well-defined grammar and spelling ("this word is written like this", "this verb is conjugated like this")?]}\\
Seleziona una delle opzioni che meglio descrive la tua opinione.\\
\textit{[Select one of the options that best describes your opinion.]}\\
\{D'accordo, Neutrale, In disaccordo\}

\textbf{Quando le persone scrivono in piemontese usano questa grammatica?}\\
\textit{[When people write in Piedmontese, do they use this grammar?]}\\
Seleziona una delle opzioni che meglio descrive la tua opinione.\\
\textit{[Select one of the options that best describes your opinion.]}\\
\{D'accordo, Neutrale, In disaccordo\}

\textbf{Da chi hai imparato il piemontese?}\\
\textit{[Where did you learn Piedmontese from?]}\\
Puoi selezionare più opzioni. Se selezioni "Altro", puoi specificare.\\
\textit{[You can select multiple options. If you select "Other," you can specify.]}
\begin{compactitem}
    \item Nonni \textit{[Grand parents]}
    \item Genitori \textit{[Parents]}
    \item Parenti \textit{[Relatives]}
    \item Amici o colleghi  \textit{[Friends or colleagues]}
    \item Altro \textit{[Other]}
\end{compactitem}

\textbf{Qual è la tua fascia d'età?}\\
\textit{[What is your age range?]}\\
Fai 30 anni tra 4 giorni? Seleziona "Tra 20 e 30" --- Hai compiuto 40 l'altro ieri? Seleziona "Tra 40 e 50"\\
\textit{[Are you turning 30 in 4 days? Select "Between 20 and 30." --- Did you turn 40 the day before yesterday? Select "Between 40 and 50."]}
\begin{compactitem}
    \item Meno di 20 \textit{[Less than 20]}
    \item Tra 20 e 30 \textit{[Between 20 and 30]}
    \item Tra 30 e 40 \textit{[Between 30 and 40]}
    \item Tra 40 e 50 \textit{[Between 40 and 50]}
    \item Tra 50 e 60 \textit{[Between 50 and 60]}
    \item Più di 60 \textit{[More than 60]}
    \item Preferisco non rispondere \textit{[I prefer not to answer]}
\end{compactitem}
\end{tcolorbox}

\begin{tcolorbox}[colback=Green!20!white,colframe=Green!100!black,breakable,use color stack, title=\textsc{Traduzione\\\textit{[Translation]}}]
\textbf{Come scriveresti questa frase in piemontese?}\\
\textit{[How would you write this sentence in Piedmontese?]}\\
Linee guida:
\begin{compactitem}
\item Se non sai come tradurla non scrivere nulla. Se vuoi puoi riprovare e il questionario dovrebbe proporre una frase casuale diversa.
\textit{[If you don't know how to translate it, don't write anything. If you want, you can try again and the questionnaire should suggest a different random sentence.]}
\item Non usare traduttori automatici (Google Translate, ecc.).
\textit{[Do not use automatic translators (Google Translate, etc.).]}
\item Non aggiungere spiegazioni (no "la traduzione è:", "... (vuole anche dire ...)") o diverse traduzioni possibili (no: "... (che vuole dire ...)", "... opzione 1/opzione 2 ...").
\textit{[Do not add explanations (no "the translation is:", "... (also means ...)") or multiple possible translations (no: "... (which means ...)", "... option 1/option 2 ...").]}
\item Può essere che alcune parole siano difficilmente traducibili. Scrivile come le scriveresti tu.
\textit{[Some words may be difficult to translate. Write them as you would.]}
\item Puoi chiedere aiuto ai nonni.
\textit{[You can ask your grandparents for help.]}
\item Accenti e simboliche magari non sono sulla tastiera ('a' come esempio). Da telefono puoi tenere premuta una lettera per vedere le opzioni disponibili.
\textit{[Accents and symbols may not be on the keyboard ('a' as an example). On your phone, you can press and hold a letter to see the available options.]}\\
\`a: /'a, \'a: /"a, \^a: /\^{}a, \~a: \textasciitilde a, \"a: /:a, \c a: /,a, \.a: /.a, \aa: /°a,
\u a: /=a, \o: //o
\end{compactitem}

\textbf{In italiano} \textit{[In Italian]}\\
\texttt{Sample}\\
\textbf{In piemontese} \textit{[In Piedmonetese]}\\
\texttt{Text field}
\end{tcolorbox}

\begin{tcolorbox}[colback=Green!20!white,colframe=Green!100!black,breakable,use color stack, title=\textsc{Valutazione\\\textit{[Evaluation]}}]
\textbf{Come valuteresti la seguente traduzione?}\\
\textit{[How would you rate the following translation?]}
Considera possibili variazioni del piemontese (ad esempio di qualcuno di Torino o di Verduno). La traduzione è stata fatta da un'altro utente e presentata senza alcuna modifica.\\
\textit{[Consider possible variations in Piedmontese (for example, someone from Turin or Verduno). The translation was done by another user and presented unchanged.]}\\

\textbf{In italiano} \textit{[In Italian]}\\
\texttt{Sample}\\
\textbf{In piemontese} \textit{[In Piedmonetese]}\\
\texttt{Sample}\\
\begin{compactitem}
\item Interamente corretta o quasi \textit{[Completely correct or almost]}
\item Probabilmente corretta, l'avrei scritta in altro modo \textit{[Probably correct, I would have written it differently]}
\item Parzialmente corretta \textit{[Partially correct]}
\item Totalmente sbagliata o quasi \textit{[Totally wrong or almost]}
\item Non lo so \textit{[I do not know]}
\item Risposta mancante, offensiva o non pertinente \textit{[Missing, offensive or irrelevant response]}
\end{compactitem}
\end{tcolorbox}

\twocolumn

\end{document}